%% file: main.tex
\documentclass[sigconf]{aamas}

\usepackage{balance} %

\setcopyright{ifaamas}
\acmConference[AAMAS '21]{Proc.\@ of the 20th International Conference on Autonomous Agents and Multiagent Systems (AAMAS 2021)}{May 3--7, 2021}{London, UK}{U.~Endriss, A.~Now\'{e}, F.~Dignum, A.~Lomuscio (eds.)}
\copyrightyear{2021}
\acmYear{2021}
\acmDOI{}
\acmPrice{}
\acmISBN{}

\acmSubmissionID{266}

\usepackage{hyperref}       %
\usepackage{booktabs}       %
\usepackage{amsfonts,amsmath}       %
\usepackage{nicefrac}       %
\usepackage{microtype}      %
\usepackage{subcaption}
\usepackage{cleveref}
\usepackage{siunitx}

\usepackage[noend]{algpseudocode}

\usepackage{algorithm,algorithmicx}
\usepackage{appendix}
\usepackage{tikz}
\usepackage{todonotes}
\usepackage{doi}
\usepackage{wrapfig}
\usepackage{natbib}

\makeatletter
\newcommand{\printfnsymbol}[1]{%
  \textsuperscript{\@fnsymbol{#1}}%
}
\makeatother

\input{sysadmin_preamble}

\title{Scalable Anytime Planning for Multi-Agent MDPs}

\author{Shushman Choudhury$^*$, Jayesh~K.~Gupta$^*$}\thanks{$^*$ Equal contribution.}
\affiliation{
  \institution{Stanford University}
  }
\email{{shushman,jkg}@cs.stanford.edu}

\author{Peter Morales}
\affiliation{
  \institution{Microsoft}}
\email{pmorales@microsoft.com}

\author{Mykel J. Kochenderfer}
\affiliation{\institution{Stanford University}}
\email{mykel@stanford.edu}

\begin{abstract}
  We present a scalable tree search planning algorithm for large multi-agent sequential decision problems that require dynamic collaboration.
  Teams of agents need to coordinate decisions in many domains,
  but naive approaches fail due to the exponential growth of the joint action space with the number of agents.
  We circumvent this complexity through an anytime approach that allows us to trade computation for
  approximation quality and also dynamically coordinate actions. 
  Our algorithm comprises three elements: online planning with Monte Carlo
  Tree Search (MCTS), factored representations of local agent interactions with coordination graphs, and
  the iterative Max-Plus method for joint action selection.
  We evaluate our approach on the benchmark SysAdmin domain with static coordination graphs and achieve comparable performance with much lower computation cost than our MCTS baselines.
  We also introduce a multi-drone delivery domain with dynamic, i.e., state-dependent coordination graphs, and demonstrate how our approach scales to large problems on this domain that are intractable for other MCTS methods. 
  We provide an open-source implementation of our algorithm at \url{https://github.com/JuliaPOMDP/FactoredValueMCTS.jl}.
\end{abstract}

\begin{document}

\pagestyle{fancy}
\fancyhead{}

\maketitle

\input{introduction}

\input{related}

\input{mcts-maxplus}

\input{results}

\input{conclusion}
\begin{acks}

This research is supported by the Ford Motor Company and the Under Secretary of Defense for Research and Engineering under Air Force Contract No. FA8702-15-D-0001. Any opinions, findings, conclusions or recommendations expressed in this material are those of the authors and do not necessarily reflect the views of the Under Secretary of Defense for Research and Engineering.

\end{acks}

\bibliographystyle{ACM-Reference-Format}
\bibliography{references}

\clearpage
\input{appendix}

\end{document}

%% file: sysadmin_preamble.tex
\usetikzlibrary{fit}
\tikzset{
  comp/.style = {
    minimum width  = 8cm,
    minimum height = 4.5cm,
    text width     = 8cm,
    inner sep      = 0pt,
    text           = white,
    align          = center,
    font           = \Huge,
    transform shape,
    thick
  },
  monitor/.style = {draw = none, xscale = 18/16, yscale = 11/9},
  display/.style = {shading = axis, left color = black!60, right color = black},
  ut/.style      = {fill = gray}
}
\tikzset{
  computer/.pic = {
    \node(-m) [comp, pic actions, monitor]
    {\phantom{\parbox{\linewidth}{\tikzpictext}}};
    \node[comp, pic actions, display] {\tikzpictext};
    \begin{scope}[x = (-m.east), y = (-m.north)]
      \path[pic actions, draw = none]
      ([yshift=2\pgflinewidth]-0.1,-1) -- (-0.1,-1.3) -- (-1,-1.3) --
      (-1,-2.4) -- (1,-2.4) -- (1,-1.3) -- (0.1,-1.3) --
      ([yshift=2\pgflinewidth]0.1,-1);
      \path[ut]
      (-1,-2.4) rectangle (1,-1.3)
      (-0.9,-1.4) -- (-0.7,-2.3) -- (0.7,-2.3) -- (0.9,-1.4) -- cycle;
      \path[pic actions, fill = none]
      (-1,1) -- (-1,-1) -- (-0.1,-1) -- (-0.1,-1.3) -- (-1,-1.3) --
      (-1,-2.4) coordinate(sw)coordinate[pos=0.5] (-b west) --
      (1,-2.4) -- (1,-1.3) coordinate[pos=0.5] (-b east) --
      (0.1,-1.3) -- (0.1,-1) -- (1,-1) -- (1,1) -- cycle;
      \node(-c) [fit = (sw)(-m.north east), inner sep = 0pt] {};
    \end{scope}
  }
}

\definecolor{turquoise}{RGB}{26,188,156}
\definecolor{emerland}{RGB}{46,204,113}
\definecolor{peterriver}{RGB}{52,152,219}
\definecolor{amethyst}{RGB}{155,89,182}
\definecolor{wetasphalt}{RGB}{52,73,94}
\definecolor{greensea}{RGB}{22,160,133}
\definecolor{nephritis}{RGB}{22,160,133}
\definecolor{belizehole}{RGB}{41,128,185}
\definecolor{wisteria}{RGB}{142,68,173}
\definecolor{midnightblue}{RGB}{44,62,80}
\definecolor{sunflower}{RGB}{241,196,15}
\definecolor{carrot}{RGB}{230,126,34}
\definecolor{alizarin}{RGB}{231,76,60}
\definecolor{clouds}{RGB}{236,240,241}
\definecolor{concrete}{RGB}{149,165,166}
\definecolor{orange}{RGB}{243,156,18}
\definecolor{pumpkin}{RGB}{211,84,0}
\definecolor{pomegranate}{RGB}{192,57,43}
\definecolor{silver}{RGB}{189,195,199}
\definecolor{asbestos}{RGB}{127,140,141}

%% file: introduction.tex
\section{Introduction}
\label{sec:intro}

Coordination is crucial for effective decision-making in cooperative multi-agent systems with a shared objective.
Various real-world problems like formation control~\citep{oh2015survey}, package delivery~\citep{ChoudhurySoloveyETAL2020}, and firefighting~\cite{DBLP:conf/atal/OliehoekSWV08} require a team of autonomous agents to perform a common task.
Such cooperative sequential decision-making problems can be modeled as a multi-agent Markov decision process (MMDP)~\cite{boutilier1996planning}, an extension of the Markov decision process (MDP) \cite{kochenderfer2015decision}.
MMDPs can be reduced to centralized single-agent MDPs with a joint action space for all agents.
Such reductions often make large problems intractable because the action space grows exponentially with the number of agents. Solving independent MDPs for all agents yields suboptimal behavior in problems where reasoning about the effects of joint actions is necessary for better performance~\citep{matignon2012}.

Many previous MMDP approaches have tried to balance these extremes of optimality and efficiency.
In the \emph{offline} setting, these include ad hoc function decomposition approaches, such as Value Decomposition Networks~\citep{sunehag2018value} and QMIX~\citep{rashid2018qmix}, or parameter sharing in decentralized policy optimization~\citep{gupta2017cooperative}.
\citet{Guestrin2002-il} introduced the concept of a coordination graph to reason about joint value estimates from a factored representation, while~\citet{kok2004sparse} used approximations to scale these ideas to larger problems.
Monte Carlo Tree Search (MCTS)~\citep{DBLP:journals/tciaig/BrownePWLCRTPSC12}, a common approach to \emph{online} planning, has been combined with coordination graphs in Factored Value MCTS~\citep{Amato2014-io}. 
However, Factored Value MCTS coordinates actions with an exact Variable Elimination (Var-El) step, which negates the anytime nature of MCTS planning.

The key idea of this paper is to \emph{recover the anytime nature of MCTS planning for MMDPs requiring coordination and also scale to larger teams of agents}.
To that end, we propose combining Max-Plus action selection, introduced by \citet{Vlassis2004-da}, with the factored value MCTS of \citet{Amato2014-io}. We do so for many reasons.
Unlike Var-El, which is exact, Max-Plus is an iterative procedure and allows for truly anytime behavior that can trade computation for approximation quality. The representation of Max-Plus is much more efficient than that of Var-El for using
dynamic, i.e., state-dependent, coordination graphs (state-dependent data-structures are a key benefit of online planning for MDPs). Finally, Max-Plus can scale to much larger 
and denser coordination graphs than Var-El, and it can be distributed for additional scalability~\cite{DBLP:conf/bnaic/KokV05}.

We present a scalable anytime MMDP planning algorithm called Factored Value MCTS with Max-Plus. On the standard SysAdmin benchmark domain~\citep{Guestrin2002-il}, with static coordination graphs, we demonstrate that our approach performs as well as or better than Factored Value MCTS with Var-El~\cite{Amato2014-io} and is much faster for the same tree search hyperparameters.
We also introduce a new MMDP domain, Multi-Drone Delivery, with dynamic coordination graphs. On the second domain, we show how our approach scales to problem sizes that are entirely intractable for other MCTS variants, while also achieving better performance on smaller problem sizes. 

%% file: related.tex
\section{Background and Related Work}
\label{sec:related}

We first review Markov decision processes (MDPs) and their multi-agent formulation. We then describe how coordination graphs can efficiently exploit locality of interactions in multi-agent problems. Finally, we discuss how to use coordination graphs to solve multi-agent MDPs.

\subsection{Multi-Agent Markov Decision Processes}
\label{sec:related-mmdps}

An MDP is defined by the tuple $(\mathcal{S}, \mathcal{A}, {T}, R)$, where $\mathcal{S}$ is the state space, $\mathcal{A}$ is the action space, $T: \mathcal{S} \times \mathcal{A} \times \mathcal{S} \to [0, 1]$
is the transition function, and $R: \mathcal{S} \times \mathcal{A} \to \mathbb{R}$ is the reward function.
The objective for solving an MDP is to obtain a \emph{policy}, $\pi: \mathcal{S} \times \mathcal{A} \to [0,1]$ that specifies a probability distribution over actions for the agent to take from its current state  to maximize its \emph{value}, i.e. its expected cumulative reward. An action-value function $Q(s, a)$ defines the expected cumulative reward after taking action $a$ in state $s$ before following the specified policy.

We focus on decision-making settings where multiple agents cooperate to achieve a shared objective~\cite{boutilier1996planning}. 
Such problems are multi-agent Markov decision processes (MMDPs), where each agent takes an individual action and the controller policy %
observes the states of all agents.
In principle, we can solve an MMDP as a standard MDP with a joint action space $\mathcal{A} = \prod_{i} \mathcal{A}_i$ ~\citep{DBLP:journals/jair/PynadathT02}.
There exist both offline and online methods for computing such MDP  policies~\cite{bertsekas2005dynamic}.

\emph{Offline} methods pre-compute a policy over the entire state space (exactly or approximately) and query the policy during execution. Various exact offline methods exist, but reinforcement learning has emerged as an attractive solution technique due to the complexity of planning in large MMDPs~\citep{DBLP:books/lib/SuttonB98}.
Reinforcement Learning approaches attempt to compute an effective value function $Q(s,a)$ 
or a policy $\pi(a \mid s)$
through repeated interaction with the environment model.
They still have difficulty with the size of the joint action space, which is exponential in the number of agents. A common strategy is to \emph{decentralize} the policy or value function, such that each only depends on the actions of a single agent~\citep{sunehag2018value,rashid2018qmix,gupta2017cooperative}.
Unfortunately, such decentralization approaches are often suboptimal for coordination and encounter exploration bottlenecks due to uncooperative random actions from the agents~\citep{Bohmer2019-zv}.

\emph{Online} methods use an alternative strategy to deal with the  complexity of multi-agent planning; they interleave planning and execution by focusing only on states that are reachable for the current state, while computing the next action to take. Monte Carlo Tree Search (MCTS) is the predominant framework for online planning and has succeeded in a variety of domains~\cite{DBLP:journals/tciaig/BrownePWLCRTPSC12}, including in multi-player contexts~\citep{nijssen2011,zerbel2019multiagent}.
The \emph{anytime} nature of MCTS (search depth and number of simulations) allows us to trade computation time for approximation quality.
However, the exponentially large action space of MMDPs can still be a bottleneck for the naive application of MCTS techniques~\cite{chaslot2008progressive}.
Dec-MCTS tries to work around this bottleneck but does not apply to action-dependent stochastic transitions of an MDP, as it directly chooses the next state~\citep{Best2019-rr}.

\subsection{Coordination Graphs and Variable Elimination}
\label{sec:related-cgs}

Several real-world multi-agent systems demonstrate \emph{locality of interaction}, i.e. the
outcome of an agent's action depends only on the actions of a subset of other agents.
The coordination graph (CG) structure is often used to encode such interactions~\cite{Guestrin2002-il,Guestrin2003}. A CG for a multi-agent system has one node
per agent, and edges connect agents if their payoffs depend on each other.
For now, we assume a stateless or single-shot decision setting (rather than sequential).
The CG structure induces a set of payoff components, \emph{where each component is associated with a clique}, i.e., a subset of agents that are all mutually connected.

For CGs in multi-agent settings, we assume that we can factor the global payoff for a joint action as the sum of local component payoffs, i.e. $Q(\overline{a}) = \sum_{c} Q_c(\overline{a}_c)$, where $\overline{a}$ is the global joint action, and $\overline{a}_c$ is the local component action (the projection of $\overline{a}$ corresponding to component $c$).
Given this factored representation and the local component payoffs%
, we can compute exactly the best joint action, $\mathrm{argmax}_{\overline{a}} \ Q(\overline{a})$,
with the Variable Elimination (Var-El) algorithm originating from the probabilistic inference literature~\cite{Guestrin2003}. Computing the optimal
joint action in a CG is equivalent to obtaining the maximum a posteriori configuration in an undirected probabilistic graphical model~\cite{Vlassis2004-da}.

\begin{figure}
    \centering
    \begin{subfigure}{0.49\columnwidth}
        \centering
        \includegraphics[width=0.99\textwidth]{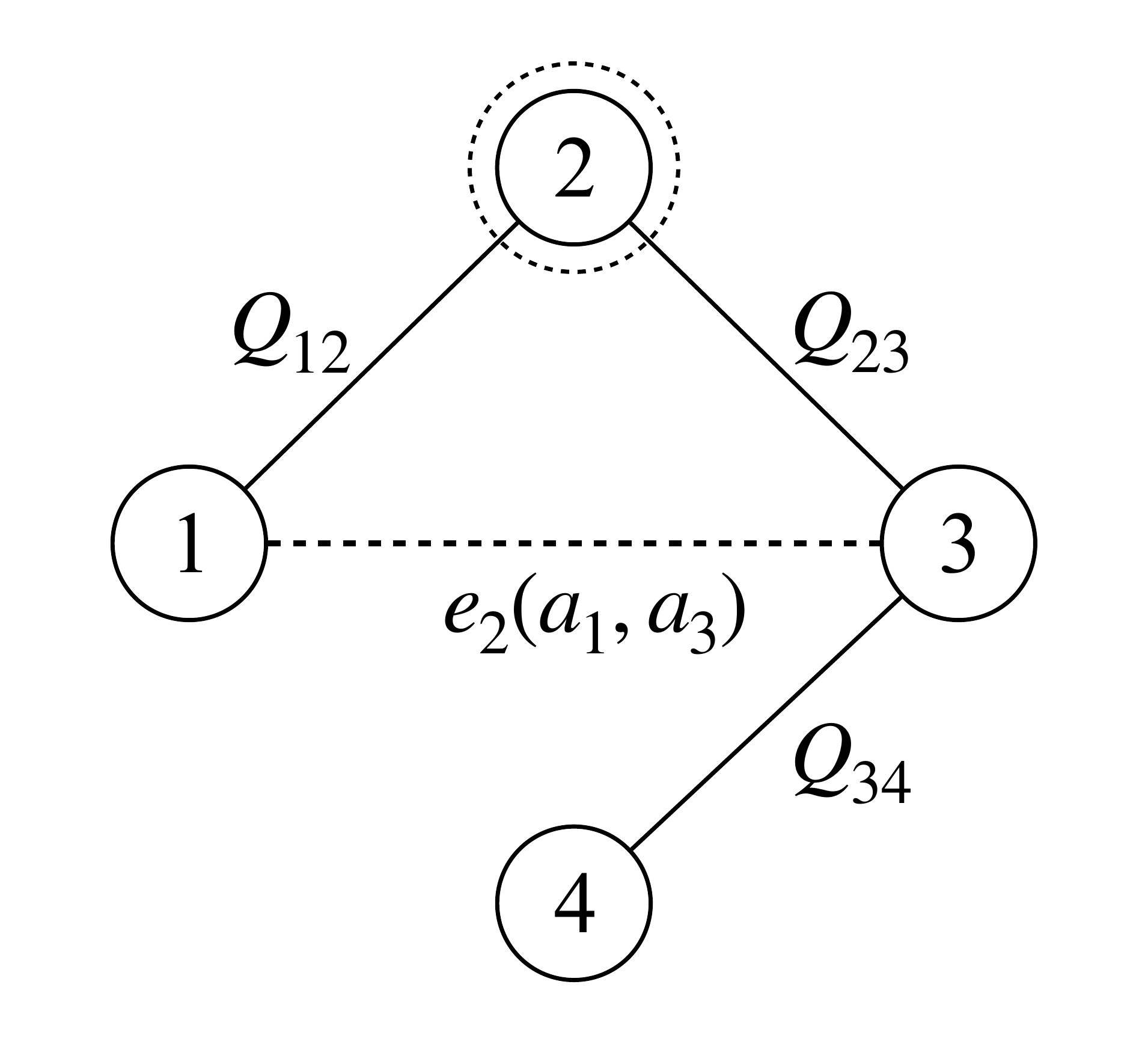}
        \caption{}
        \label{fig:cgfig-varel}
    \end{subfigure}
    \begin{subfigure}{0.49\columnwidth}
        \centering
        \includegraphics[width=0.99\textwidth]{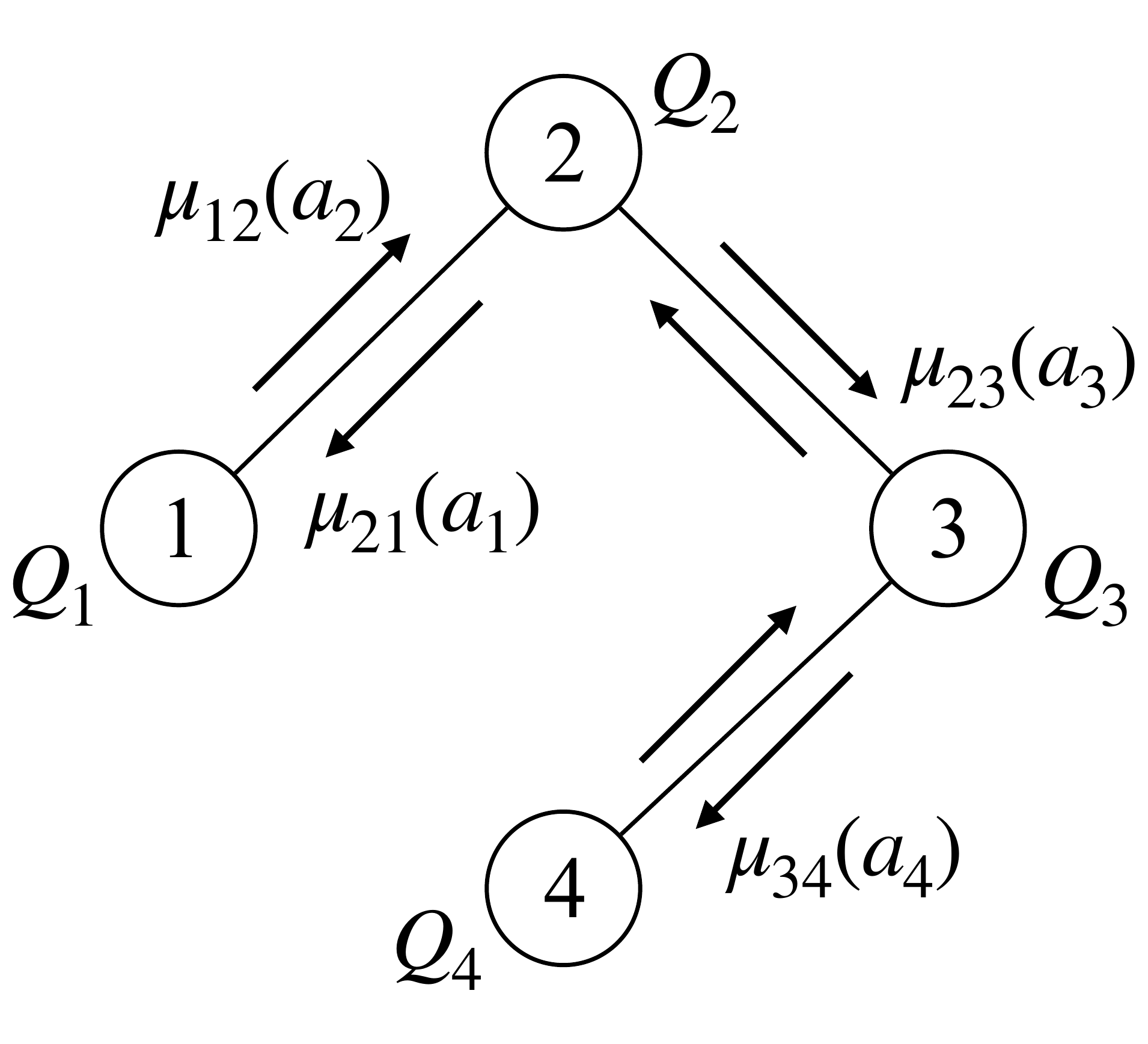}
        \caption{}
        \label{fig:cgfig-maxplus}
    \end{subfigure}
    \caption{A coordination graph for a $4$-agent MMDP with $4$ agents. (\subref{fig:cgfig-varel}) Eliminating agent $2$ in Var-El introduces an edge between nodes (agents) $1$ and $3$ and a
    new payoff function $e_2$ (\subref{fig:cgfig-maxplus}) In Max-Plus, agents passes messages along the graph edges; the messages are functions of the actions of the receiving agent, e.g.,
     agent $1$ sends $\mu_{12}(a_2)$ to agent $2$.} 
    \label{fig:cgfig}
\end{figure}

Consider the $4$-agent CG in \Cref{fig:cgfig-varel}. Here, $Q(\overline{a}) = Q_{12}(a_1,a_2) + Q_{23}(a_2,a_3) + Q_{34}(a_3,a_4)$, where $a_i$ is the action variable for agent $i$. In Var-El, we \emph{eliminate}, i.e., maximize over variables one at a time by collecting the local payoffs that depend on them. For instance, if we start with agent $2$, then
\begin{equation}
    \max_{a_1,a_3,a_4} Q_{34}(a_3, a_4) + \max_{a_2} \left[Q_{12}(a_1,a_2) + Q_{23}(a_2,a_3) \right]
\end{equation}
is the first elimination. The optimal choice for agent $4$ depends on $a_2$ and $a_3$. The internal max expression is summarized by a new intermediate payoff function $e_2(a_1,a_3) = \max_{a_2} [Q_{12}(a_1,a_2) + Q_{23}(a_2,a_3)]$ and a new edge between $1$ and $3$, after which the algorithm continues with $Q_{34}$ and $e_2$. 
After all eliminations, we recover the action for each agent by maximizing the conditional functions in reverse, finally obtaining the optimal joint action. Var-El is exponential in the induced width of the CG, which depends on the elimination order~\cite{DBLP:journals/ai/Dechter99}. 

Although most works in the literature assume a domain-dependent static coordination graph structure, some incorporate state-dependent or dynamic CGs~\citep{yu2020dist}, including learning the CG structures~\citep{kok2005utile,li2020deep}.

\subsection{Scalable MMDP Methods with Coordination Graphs}
\label{sec:related-scalable}
In the \emph{offline} context of tabular RL methods, \citet{kok2004sparse} explored action inference with predefined static coordination graphs over factorized value functions;~\citet{Bohmer2019-zv} extended these ideas to the neural network function approximation regime. We focus on \emph{anytime online} planning approaches to solving MMDPs.
\citet{Amato2014-io} provide an online planning solution by combining the idea of coordination graphs and factored values with MCTS.
Although they apply their algorithm to partially observed MDPs, the key ideas are the same for
the fully observed case.
Monte Carlo planning \emph{estimates} quantities by exploring from the current state and gathering relevant statistics through interactions with a simulated generative model of the environment~\cite{DBLP:conf/nips/SilverV10}.
These statistics typically track the average simulated reward obtained for trying an (individual or joint) action, the frequency of action attempts (for Upper Confidence Bound or UCB exploration~\cite{DBLP:conf/ecml/KocsisS06}), and the number of occurences of the (individual or joint) state.

\citet{Amato2014-io} maintain \emph{local component statistics}, i.e. the mean payoff of a local component action $\overline{a}_e$ and the number of times it was attempted in that component (they call this \emph{mixture of experts optimization}, albeit with a simple maximum likelihood estimator expert).
For instance, during tree search from the current joint state $\overline{s} \equiv \{s_i\}$ (where $s_i$ is the state of agent $i)$,
suppose the system simulates a joint action $\overline{a}$ and obtains a reward vector $\overline{r}$.
Then, in a particular CG component $e$ and the corresponding local subset of the joint action $\overline{a}_e$, they augment the local component action frequency statistic $N(\overline{s},\overline{a}_e)$ by $1$ and update the local component payoff statistic $Q_e(\overline{s}, \overline{a}_e)$ as
\begin{equation}
  \label{eq:varel-q-update}
  Q_e(\overline{s}, \overline{a}_e) \doteq Q_e(\overline{s}, \overline{a}_e) + \frac{\overline{r}_e -Q_e(\overline{s}, \overline{a}_e)}{N(\overline{s},\overline{a}_e)},
\end{equation}
which is a standard running average update.
The UCB exploration step uses the current statistics to select
joint actions, i.e.,
\begin{equation}
  \label{eq:varel-exp}
  \max_{\overline{a}} \sum_{e} U_e(\overline{s}, \overline{a}_e) = \max_{\overline{a}} \sum_{e} Q_e(\overline{s}, \overline{a}_e) + c \cdot
  \sqrt{\frac{\log{N(\overline{s})}}{N(\overline{s},\overline{a}_e)} },
\end{equation}
where $N(\overline{s})$ is the visit frequency for state $\overline{s}$.
Given these local component payoffs, i.e., the $Q_e$ functions, their method computes the best joint action at the next time-step through Variable Elimination over the CG, as in~\Cref{sec:related-cgs}.
\emph{Consequently, it loses the anytime property of MCTS because exact variable elimination cannot be stopped at an intermediate step}. Although \citet{Vlassis2004-da} explored various anytime algorithms for action selection with coordination graphs, they did not investigate their interaction with online planning algorithms like MCTS.

%% file: mcts-maxplus.tex
\section{Anytime Factored-Value Monte Carlo Tree Search}
\label{sec:approach}

We now discuss our method for anytime multi-agent MDP planning with coordination graphs, \emph{Factored-Value Monte Carlo Tree Search with Max-Plus}.
To apply the mixture of experts optimization to each node of the search tree, we must define the factored statistics to maintain for each node.
Given a potentially state-dependent undirected \emph{coordination graph} (CG), $\mathcal{G} = \langle \mathcal{V}, \mathcal{E} \rangle$, we factor the CG-induced global payoff at the current state, $\overline{s}$, as follows:
\begin{equation}
  \label{eq:maxplus-payoff}
  Q(\overline{a}) = \sum_{i \in \mathcal{V}} Q_i(a_i) + \sum_{(i,j) \in \mathcal{E}} Q_{ij}(a_i,a_j).
\end{equation}
Here, $Q_{ij}$ is a local payoff function for agents $i$ and $j$ connected by edge $(i,j)$, while $Q_i$ is an individual utility function for agent $i$ (if applicable to the domain). \emph{All state-dependent quantities in this section's equations are implicitly for the current joint state $\overline{s}$}.

\begin{figure*}[t]
    \centering
    \includegraphics[width=0.9\textwidth]{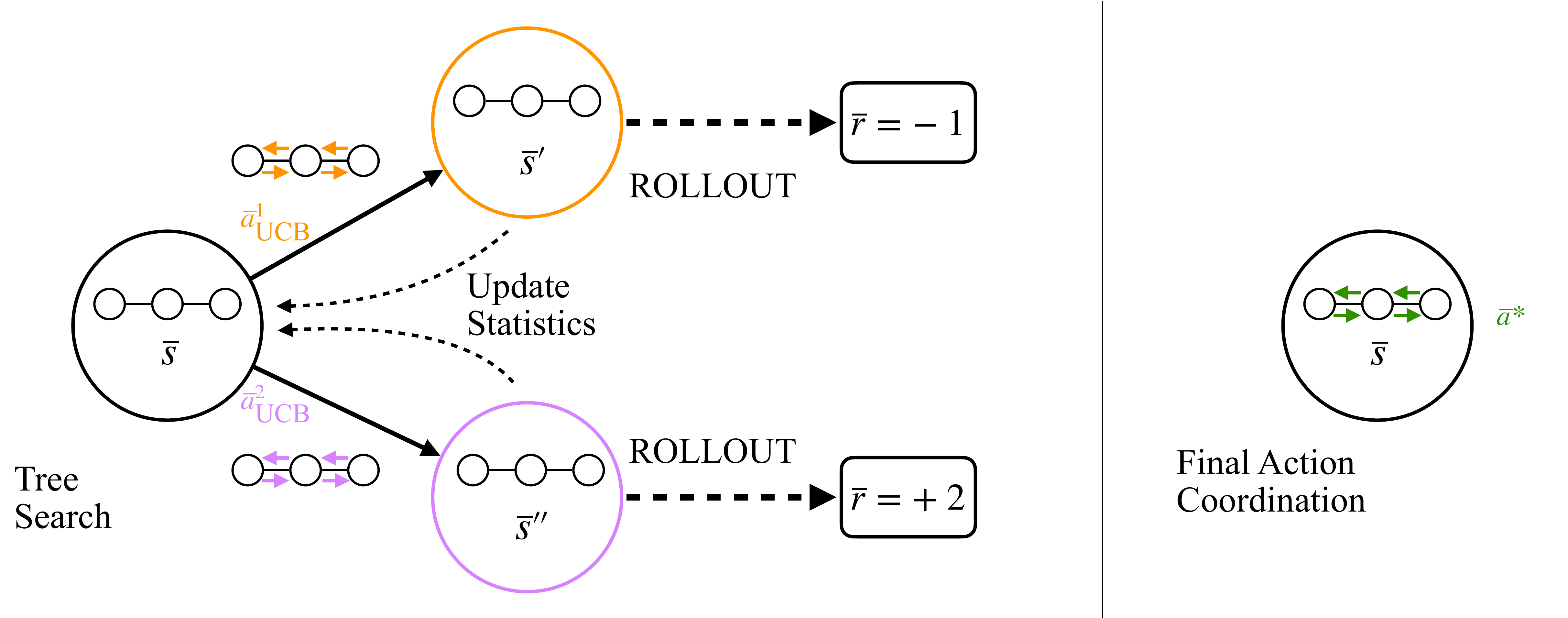}
    \caption{Our anytime MMDP planning algorithm, Factored Value MCTS with Max-Plus, computes the best joint action $\overline{a}^{*}$ for the current joint state $\overline{s}$. The tree search uses an Upper Confidence Bound (UCB) exploration bonus during action selection, while the final action coordination does not.}
    \label{fig:fv-mcts-mp}
\end{figure*}

Exploiting the duality between computing the maximum a posteriori configuration in a probabilistic graphical model and the optimal joint action in a CG, \citet{Vlassis2004-da} introduced the Max-Plus algorithm for computing the joint action via message passing. 
Each node, i.e., agent, iteratively dispatches messages to its neighbours $j \in \Gamma(i)$ in the CG (\Cref{fig:cgfig-maxplus}). A message from agent $i$ is a scalar-valued function of the action space of \emph{receiving agent} $j$, i.e.,
\begin{equation}
  \label{eq:maxplus-msg}
  \mu_{ij}(a_j) = \max_{a_i} \Big\{Q_i(a_i) + Q_{ij}(a_i,a_j) + \sum_{k \in \Gamma(i)\setminus \{j\}} \mu_{ki}(a_i)\Big\},
\end{equation}
where $\Gamma(i)$ is the set of neighbors of $i$.
Agents exchange messages until convergence or for a maximum number of rounds. 
Finally, each agent computes its optimal action individually, i.e.,
\begin{equation}
  a_i^{*} = \underset{a_i}{\mathrm{argmax}} \Big\{Q_i(a_i) + \sum_{j \in \Gamma(i)} \mu_{ji}(a_i)\Big\}
\end{equation}
Max-Plus is equivalent to belief propagation in graphical models \citep{DBLP:books/daglib/Pearl89} and its time complexity scales linearly with the CG size (the number of edges);
it is more suitable for real-time systems and more tractable for large numbers of agents than Var-El. %

Similar to Factored Value MCTS with Var-El, our method with Max-Plus (that we illustrate in~\Cref{fig:fv-mcts-mp}) is more efficient than a naive application of MCTS with the joint action space, since it retains fewer statistics and performs efficient action selection. For the rest of this section, we will discuss the key differences from the prior
work of~\citet{Amato2014-io}, which underscore how our approach is more suitable than it for large MMDPs.

\subsection{UCB Exploration with Max-Plus}
\label{sec:approach-exploration}

The key implementation issue for extending MCTS to factored value functions and coordination graphs is 
that of action exploration as per the Upper Confidence Bound (UCB) strategy. 
In the Var-El case, \citet{Amato2014-io} added the exploration bonus using component-wise statistics during each elimination step in~\Cref{eq:varel-exp}. We cannot apply this strategy with Max-Plus as it does not use components. In contrast, it has two distinct phases of computation. The first is message passing per edge in~\Cref{eq:maxplus-msg}, followed by action selection per node in~\Cref{eq:maxplus-payoff}. We use these two phases to define how our algorithm explores.

\begin{figure}[t]
    \centering
    \includegraphics[width=\columnwidth]{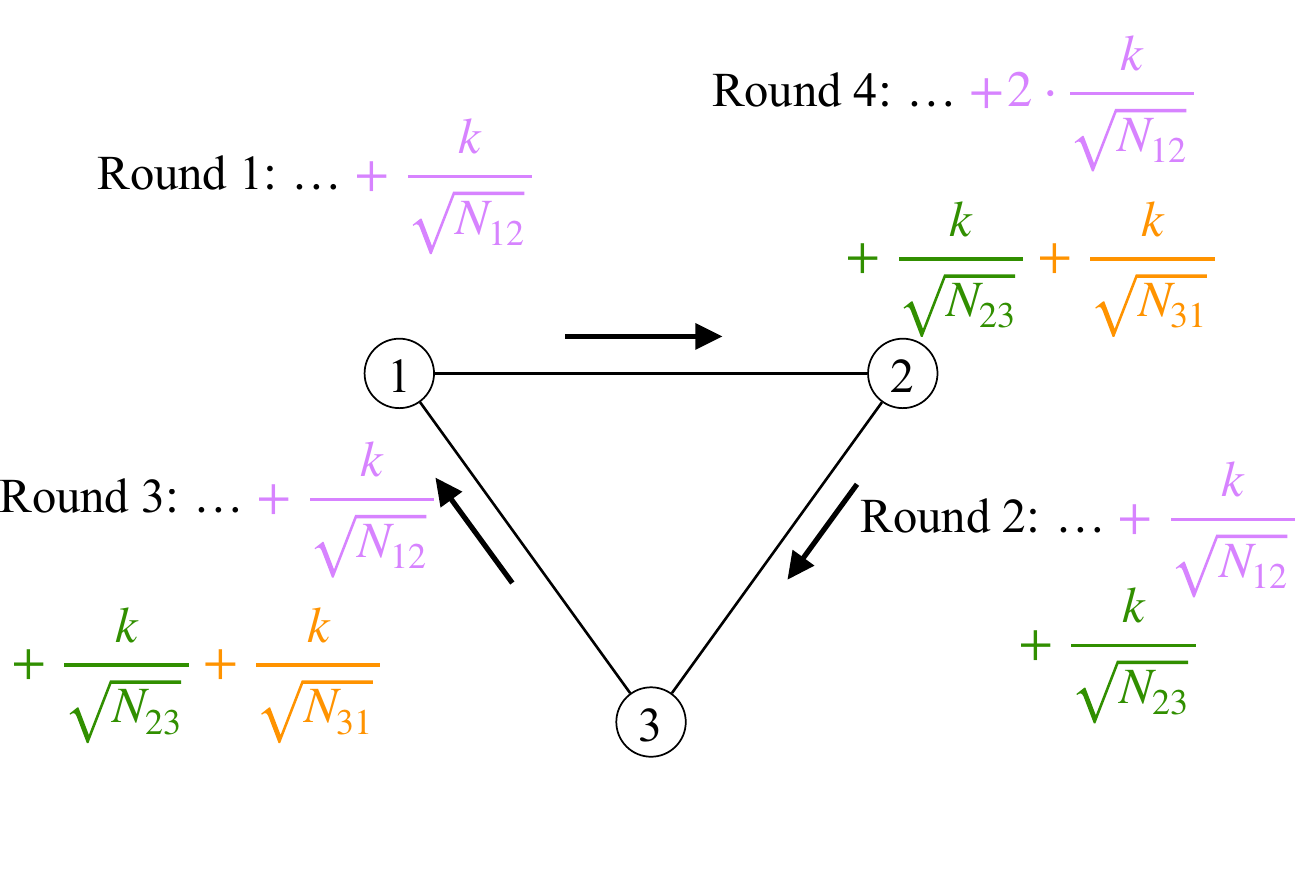}
    \caption{For coordination graphs with cycles, adding an edge exploration bonus to the messages \emph{at every round} can lead to divergent behavior. The exploration bonuses accumulate over rounds from one node to the next (clockwise or anti-clockwise along the cycle). If the number of rounds is greater than the length of a cycle (usually true), the effective relevant exploration bonus for each edge gets compounded each time the messages loop back around. We abuse some notation for convenience, i.e., $N_{ij}$ is a counting function for the pairwise actions of agents $i$ and $j$.}\label{fig:edge_exp}
    \vspace{-5pt}
\end{figure}

\textbf{Edge Exploration:} Analogous to the edge payoff statistics $Q_{ij}$, we keep track of corresponding frequency statistics $N_{a_i,a_j}$ (for pairwise actions). 
The natural exploration strategy over edges is to add the bonus to~\Cref{eq:maxplus-msg} as follows:
\begin{multline}
\label{eq:mp-edge-exp}
\mu_{ij}(a_j) = \max_{a_i} \Bigg\{Q_i(a_i) + Q_{ij}(a_i,a_j) + \\ \sum_{k \in \Gamma(i)\setminus \{j\}} \mu_{ki}(a_i) + c\sqrt{\frac{\log(N + 1)}{N_{a_i,a_j}}}\Bigg\}.
\end{multline}
Adding this bonus during the message passing rounds can cause divergence for cyclic graphs with any cycle of length less than the number of rounds.~\Cref{fig:edge_exp} illustrates intuition for this divergent behavior with a simple triangle graph. The bonuses accumulate in successive rounds for messages in either direction along the cycle, making the effective bonus proportional to the total number of rounds (divided by cycle length). Therefore, we only augment each message once \emph{after the final round of message passing}. 

\textbf{Node Exploration:} We maintain individual action frequency statistics $N_{a_i}$ and modify \Cref{eq:maxplus-payoff} to add a node exploration bonus during the action selection:
\begin{equation}
\label{eq:mp-node-exp}
  a_i^{*} = \underset{a_i}{\mathrm{argmax}} \Bigg\{Q_i(a_i) + \sum_{j \in \Gamma(i)} \mu_{ji}(a_i) +
 c\sqrt{\frac{\log(N + 1)}{N_{a_i}}}\Bigg\}    
\end{equation}

\noindent
\\
Note that the joint-action payoff $Q(\overline{a})$ can be factorized over the CG nodes and edges as in~\Cref{eq:maxplus-payoff}, but the joint-action exploration bonus $c\sqrt{\frac{\log{N(\overline{s})}}{N(\overline{s},\overline{a})}}$ cannot. Therefore, the node and edge exploration strategies we have defined here are heuristic choices that we make and will evaluate empirically through an ablation, rather than strategies we derive analytically by factorizing the exploration bonus. The same was true even in the exact action selection case of Var-El, where the analysis was restricted to the approximation quality of the factored value function, not the exploration itself. Quoting~\citet{Amato2014-io}, ``it is not possible to demonstrate that the UCB exploration policy is component-wise'', where, as from~\Cref{sec:related-cgs}, the components refer to the cliques in the CG.
\\
We outline our approach in~\Cref{algo:fvmctsmp}
as well as the Max-Plus routine in~\Cref{algo:maxplus}.

\algdef{SE}%
[STRUCT]%
{Struct}%
{EndStruct}%
[1]%
{\textbf{struct} \textsc{#1}}%
{}%

\begin{algorithm}[t]
  \caption{Factored Value MCTS with Max-Plus}
  \label{algo:fvmctsmp}
  \begin{algorithmic}[1]
    \Require{time limit, depth, exploration constant $c$, state $\overline{s}$}
    \Statex
    \State Initialize $N_i, Q_i$  \Comment{Node statistics}
    \State Initialize $N_{ij}, Q_{ij}$ \Comment{Edge statistics}
    \State
    \Function{FV-MCTS-MP}{$\overline{s}$, depth}
    \While{time limit not reached}
    \State{\Call{Simulate}{$\overline{s}$, depth}}
    \EndWhile
    \State{$\overline{a}^* \gets$ \Call{MaxPlus}{$0$}} \Comment{No exploration here}
    \State{\Return{$\overline{a}^*$}} \Comment{Best joint action}
    \EndFunction

    \Function{Simulate}{$\overline{s}$, depth}
    \If{$\mathrm{depth} = 0$}
    \State{\Return $0$}
    \EndIf
    \State{$\overline{a}\gets$ \Call{MaxPlus}{$c$}}
    \State{$\overline{s}', \overline{r} \sim T(\overline{s},{\overline{a}}), R(\overline{s},\overline{a})$} \Comment{Generative model}
    \State{$\overline{q} \gets \overline{r} + \gamma \cdot$ \Call{Simulate}{$\overline{s}'$, $\text{depth} - 1$}}
    \State{\Call{UpdateStats}{$\overline{s}, \overline{a}, \overline{q}$}}
    \EndFunction
    
    \Function{UpdateStats}{$\overline{s}, \overline{a}, \overline{q}$}
    \For{every agent $i$}
    \State{$N_i(\overline{s}, a_i) \mathrel{{+}{=}} 1$}
    \State{$Q_i(\overline{s}, a_i) \mathrel{{+}{=}} \frac{q_i - Q_i(\overline{s}, a_i)}{N_i(\overline{s}, a_i)}$}
    \EndFor
    \For{every edge $(i, j) \in \mathcal{G}(\overline{s})$}
    \State{$N_{ij}(\overline{s}, a_i, a_j) \mathrel{{+}{=}} 1$}
    \State{$q_e \gets q_i + q_j$}
    \State{$Q_{ij}(\overline{s}, a_i, a_j) \mathrel{{+}{=}} \frac{q_e - Q_{ij}(\overline{s}, a_i, a_j)}{N_{ij}(\overline{s}, a_i, a_j)} $}
    \EndFor
    \EndFunction
  \end{algorithmic}
\end{algorithm}

\algnewcommand{\LineComment}[1]{\State \(\triangleright\) #1}

\begin{algorithm}[t]
  \caption{MaxPlus Action Selection}
  \label{algo:maxplus}
  \begin{algorithmic}[1]
    \Require{Coordination Graph $\mathcal{G}(s) = \langle \mathcal{V}, \mathcal{E}\rangle$; state node statistics $N, Q$; max iterations $M$; flags (exploration; normalization)}
    \Statex
    \Function{MaxPlus}{c}
        \For{$t \gets 1$ to $M$}
            \State{$\mu_{ij}(a_j) = \mu_{ji}= 0$ for $(i,j) \in \mathcal{E}, a_i \in \mathcal{A}_i, a_j \in \mathcal{A}_j$}
            \For{every agent $i$}
                \For{all neighbors $j \in \Gamma(i)$}
                \State{Compute $\mu_{ij}(a_j)$ via \Cref{eq:maxplus-msg}}
                \If{message normalization}
                \State{$\mu_{ij}(a_j) \mathrel{{-}{=}} \frac{1}{\lvert \mathcal{A}_j \rvert} \sum_{a_j \in \mathcal{A_j}} \mu_{ij}(a_j)$}
                \EndIf
                \State{send message $\mu_{ij}(a_j)$ to agent $j$}
                \If{$\mu_{ij}(a_j)$ close to previous message} 
                \State{break}
                \EndIf
                \EndFor
            \EndFor
            \For{every agent $i$}
                \If{edge exploration}
                \For{all neighbors $j \in \Gamma(i)$}
                \State{Compute $\mu_{ij}(a_j)$ via \Cref{eq:mp-edge-exp}}
                \EndFor
                \EndIf
                \State{$q_i(a_i) = Q_i(a_i) + \sum_{j\in \Gamma(i)} \mu_{ji}(a_i)$}                
                \If{node exploration}
                \State{$q_i(a_i) \mathrel{{+}{=}} c \sqrt{\frac{\log (N + 1)}{N_i(a_i)}}$} %
                \EndIf
                \State{$a'_{i} = \arg\max_{a_i} q_i(a_i)$}
            \EndFor
            \If{time limit reached}
            \State{break}
            \EndIf            
        \EndFor
        \State{\Return{$\overline{a}'$}}
    \EndFunction
  \end{algorithmic}
\end{algorithm}

\subsection{Other differences from FV-MCTS with Variable Elimination}
\label{sec:approach-differences}

\textbf{Convergence:}
For graphs without cycles, Max-Plus converges to a fixed point in finitely many iterations~\cite{DBLP:books/daglib/Pearl89}. 
For cyclic graphs, there are no such guarantees in general~\cite{DBLP:journals/sac/WainwrightJW04}.
However, cyclic message passing can work well in practice~\cite{DBLP:conf/uai/MurphyWJ99}.

\noindent\textbf{Agent Utilities:} The Max-Plus global payoff in~\Cref{eq:maxplus-payoff} includes a utility function $Q_i$ for each individual agent.
The FV-MCTS with Var-El has no such individual utility (unless a node has degree $0$ in the CG).
If such agent utilities were known or learned \emph{independent of the payoffs},
we would naturally use them during action coordination. 
However, in FV-MCTS we estimate all statistics from the rewards obtained during tree search with a simulated environment model; the environment model returns precisely one reward vector for each joint state-action pair. 

We already account for the simulated rewards in tree search through the $Q_{ij}$ local payoff statistics in~\Cref{eq:varel-q-update}. We do not receive independent per-agent rewards, so
utility statistics would be derived from the same information we use for the payoff statistics.
Our experiments compare the benefit of these derived individual agent (node) utilities, in addition to local edge payoffs. We maintain separate statistics $N_i$ and $Q_i$ for the per-agent frequencies and utilities respectively and estimate them from the joint rewards during tree search; the corresponding updates are  $N_i(\overline{s},\overline{a}_i) = N_i(\overline{s},\overline{a}_i) + 1$ and $Q_i(\overline{s}, \overline{a}_i) = Q_i(\overline{s}, \overline{a}_i) + \frac{\overline{r}_i - Q_i(\overline{s}, \overline{a}_i)}{N_{\overline{a}_i}}$ for an agent $i$. The results in~\Cref{sec:results-sysadmin} demonstrate how \emph{including derived node utilities} enables better empirical performance.

\noindent\textbf{Dynamic Coordination Graphs:} Recall that MCTS (and online MDP planning in general) can use computational structures that vary with the current state.
\emph{For FV-MCTS with Var-El, state-dependent or dynamic CGs are not feasible} because
eliminating an agent can change the intermediate CG topology during action coordination  (by adding edges).
It is not tractable to maintain statistics for the set of all possible CG components, the size of which is exponential in the number of agents~\cite{zykov1949some}. 
On the other hand, Max-Plus only maintains statistics for at most all CG edges, over which the messages are sent.
Therefore, dynamic CGs can be used seamlessly~\citep{yu2020dist}.

\noindent\textbf{Memory Complexity of Statistics}: Factored Value MCTS collects a set of frequency and payoff statistics for each unique joint state encountered during tree search.
Assume an MMDP with the same action set $\mathcal{A}$ for each agent, and a CG with $|\mathcal{V}|$
nodes (agents), $|\mathcal{E}|$ edges, and $C$ local components or cliques. Then, the memory complexity of per-state statistics for Max-Plus is $\mathcal{O}(|\mathcal{V}| |\mathcal{A}| + |\mathcal{E}||\mathcal{A}|^2)$ (the first term only applies if we track per-node utilities).
In contrast, the per-state memory for Var-El statistics is $\mathcal{O}(\sum_{c \in C} |\mathcal{A}|^{|\mathcal{V}|_c} \cdot |\mathcal{V}|_c)$, where $|\mathcal{V}|_c$ is the size of
local component $c$. For a CG that is connected (typically the case), the memory complexity for Var-El is at least $\mathcal{O}(|\mathcal{E}||\mathcal{A}|^2)$, which is the dominant term for Max-Plus, and more generally is exponential in the largest clique. \emph{Therefore, Max-Plus is more memory-efficient than Var-El}. The experiments in~\Cref{sec:results-drone} empirically support this claim by showing how our algorithm can solve problems that cause out-of-memory
issues for the Var-El baseline.

\noindent\textbf{Distributed Implementation:} Unlike with Var-El, \emph{we can execute Max-Plus in a distributed manner} by sending messages in parallel, albeit incurring additional communication complexity. 
Such an implementation can allow further scalability with available compute. Note that this is distinct from full decentralization wherein the agent actions can be computed independently. %

\begin{figure*}[t]
    \centering
    \begin{subfigure}{0.4\textwidth}
        \centering
        \fbox{\resizebox{0.75\columnwidth}{!}{\input{figures/star_sysadmin.tex}}}
        \caption{}
        \label{fig:domains-sysadmin}
    \end{subfigure}
    \begin{subfigure}{0.4\textwidth}
        \centering
        \fbox{\includegraphics[width=0.95\textwidth]{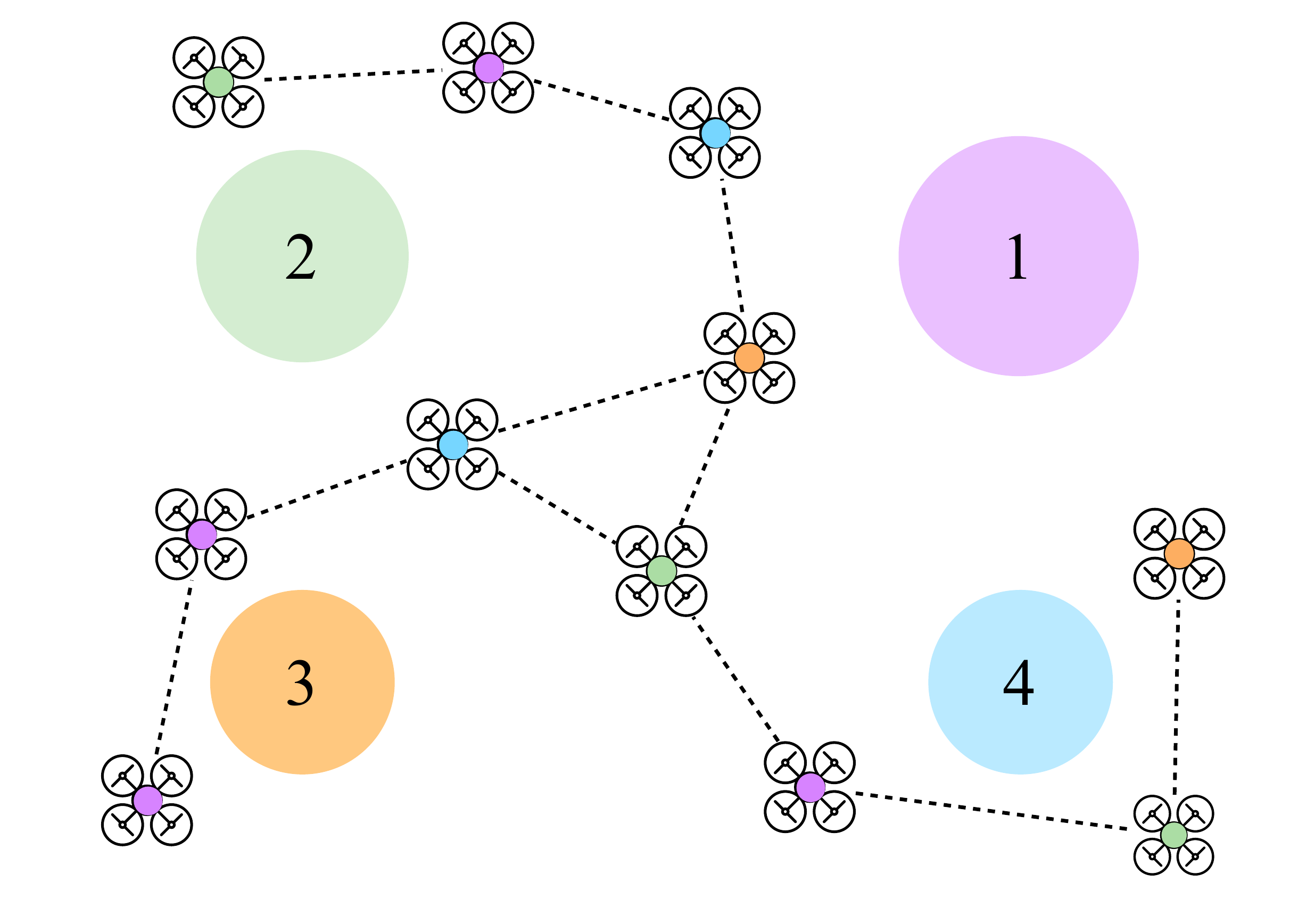}}
        \caption{}
        \label{fig:domains-multidrone}
    \end{subfigure}
    \caption{Our two experimental domains: (\subref{fig:domains-sysadmin}) SysAdmin (star topology) and (\subref{fig:domains-multidrone}) Multi-Drone Delivery, where dotted lines illustrate
    a subset of the Coordination Graph edges for the current state (for clarity, we omit some edges between drones of the same color, i.e., assigned to the same goal).}
    \label{fig:domains}
\end{figure*}

%% file: figures/star_sysadmin.tex
\begin{tikzpicture}
  \pic(comp0) [
  draw,
  fill = gray!30,
  scale = 0.4,
  pic text = {0}
  ]
  {computer};
  \path(comp0-c.center) pic
  foreach[count=\i] \farbe in {sunflower, carrot, pomegranate, peterriver, midnightblue, emerland}
  (comp\i) [
  draw = \farbe,
  fill = \farbe!30,
  scale = 0.4,
  pic text = {\i}
  ] at +(60*\i:8){computer};
  \foreach \i in {1,2,4,5} \draw (comp\i-c) -- (comp0-c);
  \foreach \i in {3,6} \draw (comp\i-m) -- (comp0-c);
\end{tikzpicture}

%% file: results.tex
\section{Experiments and Results}
\label{sec:results}

We used cumulative discounted return as the primary metric to evaluate our approach, Factored Value MCTS with Max-Plus (FV-MCTS-MP). Our most relevant baseline is Factored Value MCTS with Variable Elimination (FV-MCTS-Var-El). We also compared against standard MCTS (with no factorization), independent Q-learning (IQL), and a random policy. Besides measuring performance, we examined the effect of different exploration schemes on the performance of FV-MCTS-MP (as we discussed in~\Cref{sec:approach-differences}) and the problem size on MCTS computation time. The appendix provides performance results for FV-MCTS (both variants) with different hyperparameters. Both of our experimental domains represent a range of MMDP problems and underlying coordination graphs (CGs). The source-code for experiments is available at \url{https://sites.google.com/stanford.edu/fvmcts/}. All implementations and simulations are in Julia~\citep{bezanson2017julia}.

We will show qualitatively how our approach recovers the true anytime nature of MCTS by using Max-Plus rather than Var-El. However, \textit{there are many confounds for quantitatively evaluating the anytime property}. Our metric is the average discounted return over the episode, where the Max-Plus routine is called several times; typical anytime evaluation reports improving solution quality with more compute time for a single call to a method. MCTS itself has several parameters that affect the computation-vs-quality tradeoff, such as tree depth, exploration constant, and number of trials. With dynamic CGs as in our multi-drone delivery domain, the same Max-Plus parameters leads to different computation times. Note that our reference for Max-Plus does evaluate its anytime property in a one-shot decision-making domain that does not have any of the above confounds~\cite{Vlassis2004-da}.

\subsection{SysAdmin Domain}
\label{sec:results-sysadmin}
Our first domain is a standard MMDP benchmark: SysAdmin \citep{Guestrin2003}.
Each agent $i$ represents a machine in a network with two state
variables: Status $S_i \in \{\textsc{good}, \textsc{faulty}, \textsc{dead}\}$, and Load $L_i \in \{\textsc{idle}, \allowbreak \textsc{loaded}, \textsc{success}\}$.
A \textsc{dead} machine increases the probability that its neighbor also dies.
The system gets a reward of $1$ if a process terminates successfully,
processes take longer when status is \textsc{faulty}, and a \textsc{dead} machine loses the process. 
Each agent must decide whether to reboot its machine, in which case the Status becomes \textsc{good} and any running process is lost. 
The discount factor, $\gamma$ used in all the experiments is $0.9$.
All evaluations have been averaged over $40$ runs. Error bars indicate standard deviations.
\Cref{fig:domains-sysadmin} illustrates the star network topology for SysAdmin; we also use the ring topology as well as the ring-of-rings topology.

\begin{figure*}[t]
    \centering
    \begin{subfigure}{0.49\textwidth}
    \centering
    \includegraphics[width=0.75\columnwidth]{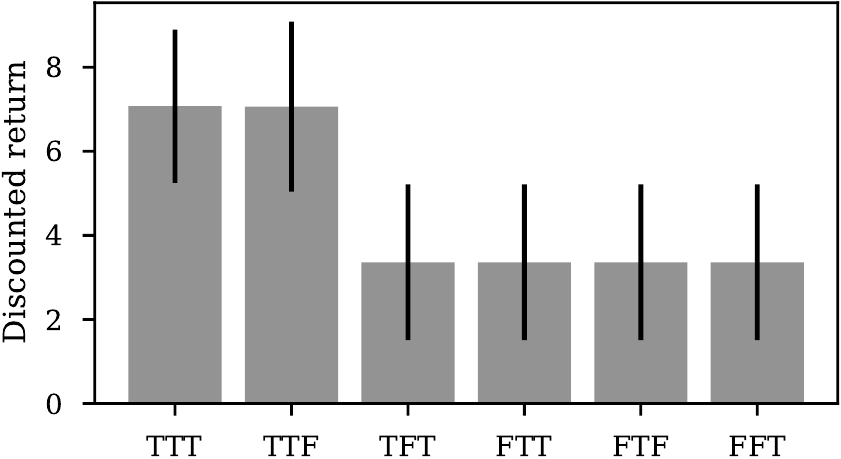}
    \end{subfigure}
    \begin{subfigure}{0.49\textwidth}
    \centering
    \includegraphics[width=0.75\columnwidth]{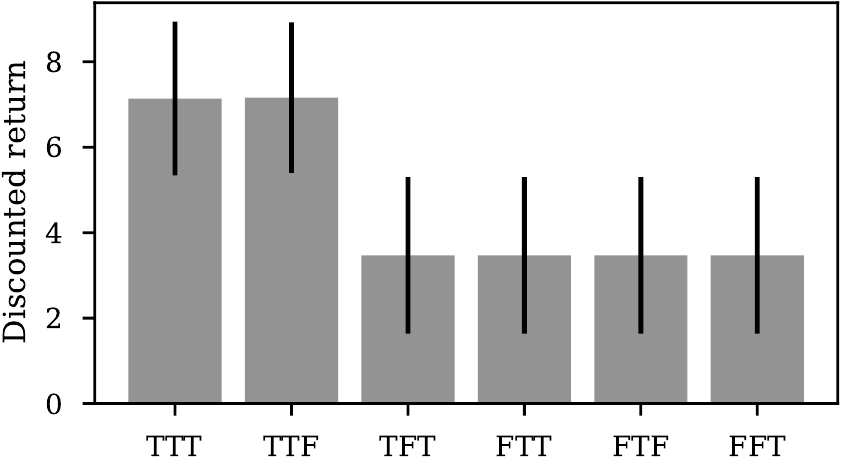}
    \end{subfigure}
    \caption{The performance of FV-MCTS-MP varies with different combinations of exploration strategies for the 4-agent Sysadmin on Ring (left) and Star (right) topologies. The True/False (T/F) labels correspond to Agent Utilities, Node Exploration and Edge Exploration in order, e.g. TTF implies agent utilities and only node (but not edge) exploration.}
    \label{fig:exploration}
\end{figure*}

\noindent
\textbf{Exploration Schemes for FV-MCTS-MP}:
The three knobs affecting exploration in FV-MCTS-MP are per-agent utility, node bonus, and edge bonus. We compared the discounted return of variants that either use or ignore per-agent utilities, and use either or both bonuses.~\Cref{fig:exploration} demonstrates that the combination of \emph{agent utilities and only node exploration (TTF) is enough}; including the edge bonus as well (TTT) does not have much effect. All other schemes are significantly poorer. Therefore, we used the TTF variant of FV-MCTS-MP to compare against the other baselines. Lack of significant difference between some of the exploration strategies has more to do with the small action space of the SysAdmin domain. %

\begin{figure*}[t]
    \centering
    \begin{subfigure}{0.33\textwidth}
    \includegraphics[width=0.99\columnwidth]{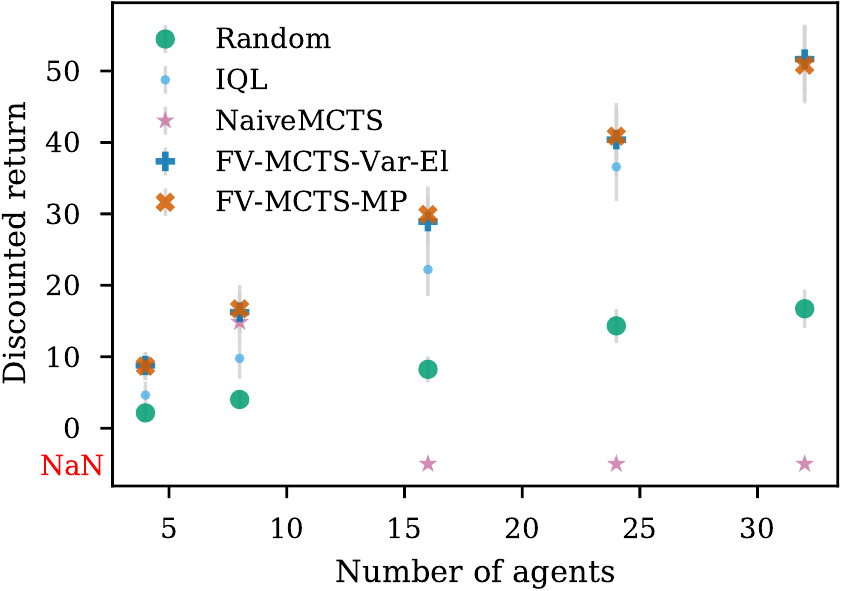}
    \end{subfigure}
    \begin{subfigure}{0.33\textwidth}
    \includegraphics[width=0.99\columnwidth]{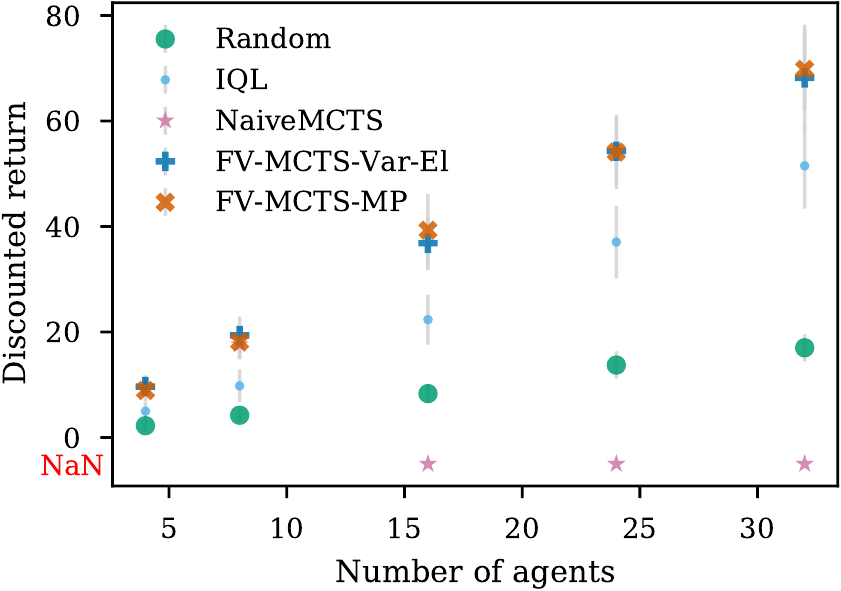}
    \end{subfigure}
    \begin{subfigure}{0.33\textwidth}
    \includegraphics[width=0.99\columnwidth]{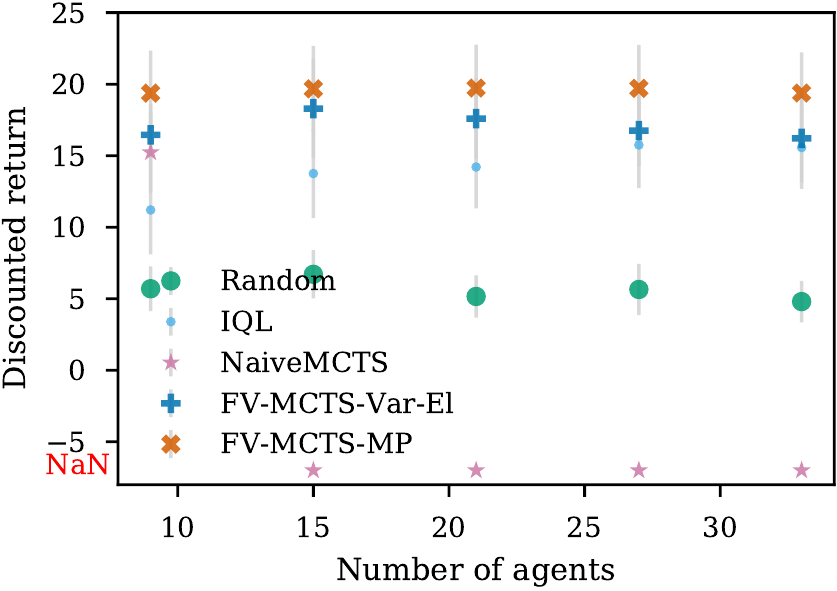}
    \end{subfigure}
    \caption{On SysAdmin topologies: Ring (left), Star (middle), and Ring-of-Rings (right), FV-MCTS with MaxPlus performs as well as or slightly better than Var-El, while being much more efficient for larger problems as in~\Cref{fig:sysadmin_timing}. \textcolor{red}{NaN} indicates that the algorithm ran out of memory.}
    \label{fig:probsize_nodes}
\end{figure*}

\noindent
\textbf{FV-MCTS-MP compared to baselines}:
For all three SysAdmin topologies (and corresponding fixed CGs), we varied the number of machines (agents) and compared the performance of all methods in~\Cref{fig:probsize_nodes}.
With fewer agents, all MCTS methods perform similarly to each other and better than Q-Learning.
However, with more agents, standard MCTS runs out of memory even on our 128GB RAM machine, as expected for large joint action spaces.
Both Factored Value MCTS variants perform comparably on larger problems (on ring-of-rings our Max-Plus variant was better). However, as we discuss subsequently, FV-MCTS-Var-El
is much slower than FV-MCTS-MP, e.g., taking approximately \SI{35}{\second}
versus \SI{16}{\second} for $32$ agents on a single-threaded implementation in the Ring topology.
\emph{Therefore our approach strictly dominates the Var-El baseline on the performance-time tradeoff}.

\noindent\textbf{Effect of Hyperparameters}:
We performed ablation experiments, varying one of exploration constant $c$, tree search exploration depth $d$ and number of Monte Carlo rollouts $n$, while keeping the rest of the hyperparameters constant.
Low values of $n$ adversely effected the performance a little for both FV-MCTS-Var-El and FV-MCTS-MP, while \emph{low values of $c$ vastly degraded the performance of FV-MCTS-Var-El only}.
The difference in performance was not significant over a range of values of $d$. 
\Cref{fig:hyperparam_fig} in the appendix, shows results for $32$ agents. Similar results hold for the other domains as well as different numbers of agents.

\begin{figure*}[t]
    \centering
    \begin{subfigure}{0.46\textwidth}
    \centering
        \includegraphics[width=0.9\columnwidth]{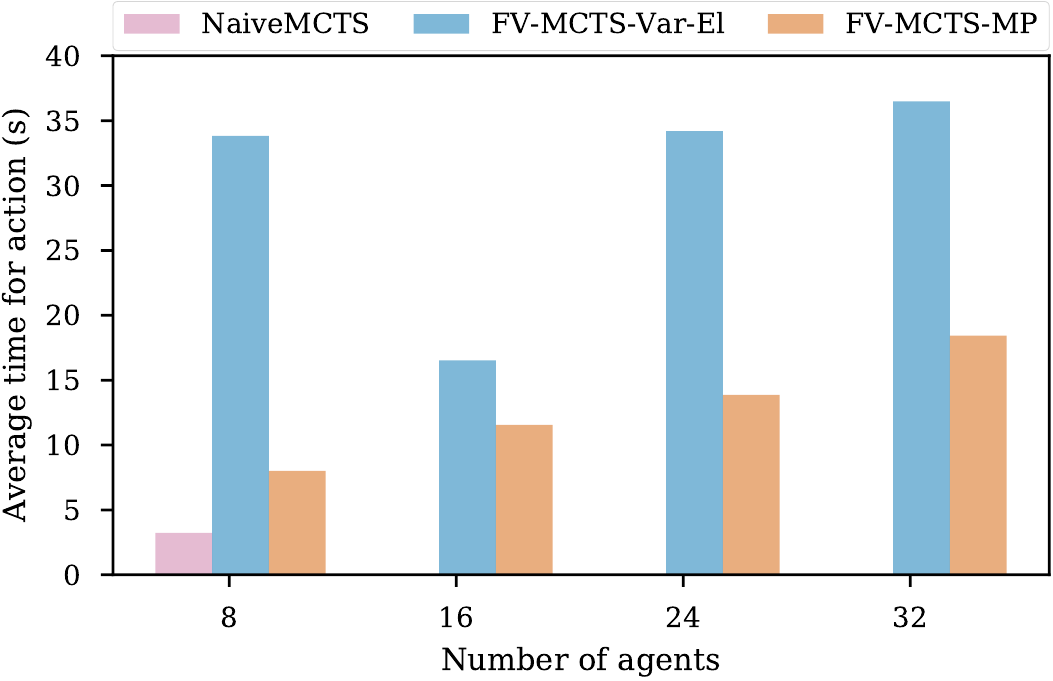}
    \end{subfigure}
    \begin{subfigure}{0.46\textwidth}
    \centering
        \includegraphics[width=0.9\columnwidth]{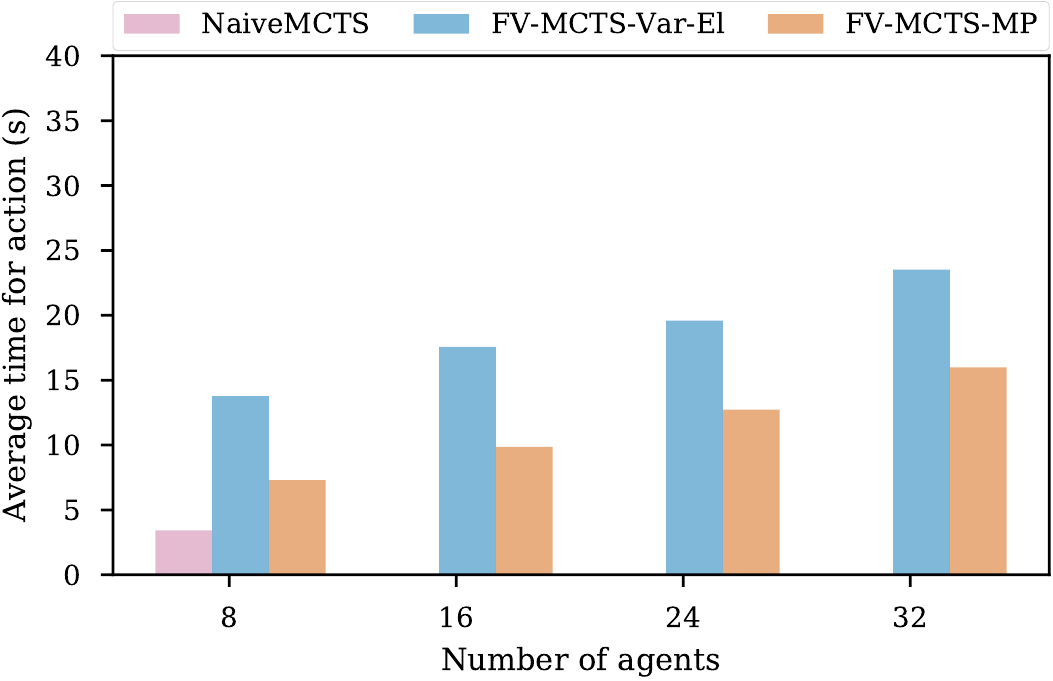}
    \end{subfigure}
    \caption{Runtime comparisons (lower is better) for the same tree search hyperparameters on SysAdmin with Ring (left) and Star (right) topologies. The NaiveMCTS baseline ran out of memory with more than $8$ agents.}
    \label{fig:sysadmin_timing}
\end{figure*}

\noindent\textbf{Computation Time}:
For the same tree search hyperparameters with number of iterations fixed as $16000$, exploration constant as $20$ and tree search depth as $20$, we compared the average time taken for each action for different number of agents in the coordination graphs.
For a fair comparison, we used a single threaded implementation.
As demonstrated in \Cref{fig:sysadmin_timing}, we found FV-MCTS-MP to be consistently faster than FV-MCTS-Var-El.
Although MCTS was faster when there were small number of agents, it ran out of memory as the number of agents increased.

\subsection{Multi-Drone Delivery Domain}
\label{sec:results-drone}
Besides the SysAdmin domain, previous multi-agent decision-making work has also used the Firefighter~\citep{Amato2014-io} and Traffic Control~\citep{kuyer2008multiagent} domains for benchmarking.
Underneath the differing high-level descriptions, however, \emph{the MMDP details of all three domains are very similar}:
a small state space and binary action space, the degrees of most nodes in the coordination graph are independent of the total number
of agents (except the hub node for Star SysAdmin), and there is no scope in any of them for dynamic CGs.

We introduce and use a truly distinct domain for our second set of experiments.
It simulates \emph{a team of delivery drones navigating a shared operation space to reach their assigned goal regions}.
We are motivated by recent advances in drone delivery technology, from high-level routing to low-level control~\cite{dorling2016vehicle,DBLP:conf/syscon/Lee17};
in particular, drones using ground vehicles as temporary modes of transit to save energy and increase effective travel range~\cite{choudhury2019dynamic,ChoudhurySoloveyETAL2020}.
Our domain models a key component of such drone-transit coordination: multiple drones assigned to board transit vehicles in close
proximity to each other (within the same time window).
\\
\noindent
\textbf{Domain details:}
\Cref{fig:domains-multidrone} illustrates our Multi-Drone Delivery domain; for convenience and consistency with MMDP benchmarks we discretize everything,
but MCTS could accommodate a continuous state space. Each drone starts in a randomly sampled unique cell in a grid (we use larger
grids for more drones in our simulations). There are four circular goal regions, one in each quadrant, that represent a transit vehicle; each
goal region has a radius and maximum capacity of drones it can accommodate, since no two drones can occupy the same grid cell (we also vary
goal radius with grid size). We allocate the drones to the goal regions at random such that at least two drones target every region.
\begin{figure}
    \centering
    \includegraphics[width=0.99\columnwidth]{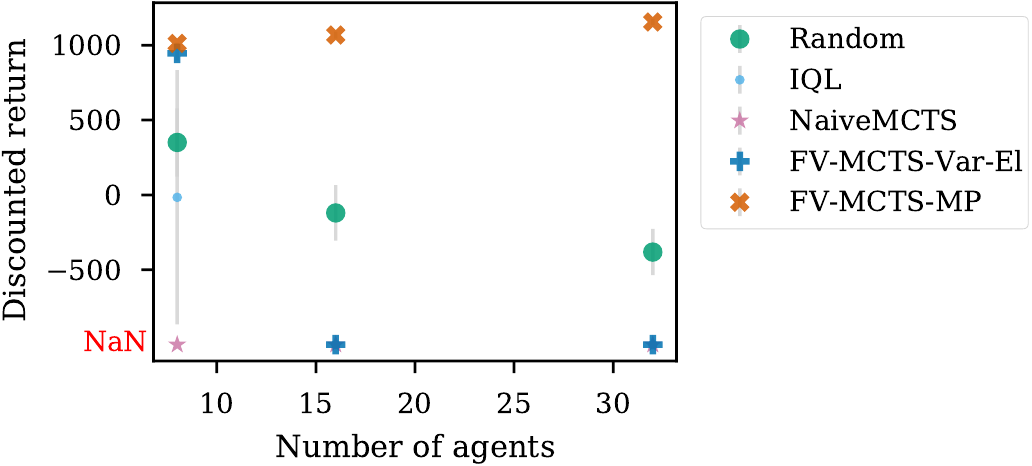}
    \caption{For Multi-Drone Delivery, FV-MCTS-MP vastly outperforms the baselines while effectively using dynamic CGs without any memory issues. \textcolor{red}{NaN} indicates that the algorithm ran out of memory.}
    \vspace{-10pt}
    \label{fig:drone_scaling}
\end{figure}

Each drone has $10$ actions in total: one for moving to each of the $8$-connected grid neighbors, a \textsc{no-op} action for staying in place,
and a \textsc{board} action that is only valid when the drone is inside its assigned goal region. The MMDP is episodic and terminates only when all drones have reached their goals and executed $\textsc{Board}$ inside them, thus boarding the transit vehicle and receiving a reward of $1000$. Drones also receive an intermediate positive reward if they get closer to their assigned goals. Besides drone movement, the other sources
of negative reward, i.e., cost, are penalties for two or more drones being too close to each other, attempting to enter the same cell (which makes them both stay in place),
and attempting to board in the same goal region at the same time.

Unlike the typical MMDP domains used in prior work, \emph{Multi-Drone Delivery motivates dynamic or state-dependent coordination graphs}; 
any two drones benefit from coordination only when they are close to each other. Therefore, at the current joint state,
we assign a CG edge between any two drones whose mutual distance is lower than a resolution-dependent threshold 
(depicted in \Cref{fig:domains-multidrone}). We also add edges apriori between all drones assigned to
the same goal region, as they need to coordinate while boarding.

For all experiments, we set the discount factor $\gamma$ to $1$, the
goal reaching reward to $1000.0$ units, and the collision penalty to $10.0$ units.~\Cref{tab:multidrone_exp} describes the full set of varying problem resolutions and MCTS hyperparameters.
The average degree of the dynamic coordination graphs ranged from $2.4$ for $8$ agents to $11.8$ for $48$ agents.
We averaged all evaluations over $20$ runs; error bars indicate standard deviations.

\begin{table}
    \centering
    \resizebox{\columnwidth}{!}{\begin{tabular}{rrrrrr}
    \toprule
    Agents    &  XY axis res. & Noise & Expl.\ const. & Expl.\ depth & Iter\\
    \midrule
    8  & 0.20  & 0.10 & 5 & 10 & 4000 \\ 
    16 & 0.10  & 0.05 & 10 & 10 & 8000 \\
    32 & 0.08 & 0.05 & 20 & 10 & 16000 \\
    48 & 0.05 & 0.02 & 30 & 10 & 24000 \\
    \bottomrule
    \end{tabular}}
    \caption{Multi-Drone Delivery hyperparameters.}
    \label{tab:multidrone_exp}
    \vspace{-10pt}
\end{table}

\noindent
\textbf{Performance of FV-MCTS-MP against Baselines}:
As with SysAdmin, we varied the number of drones (agents), discretizing the grid appropriately,
and compared against all baselines (except Random) in~\Cref{fig:drone_scaling}.
We observed that FV-MCTS-Var-El and MCTS quickly ran out of memory, which is expected given the large action space per agent. Even on the problems where Var-El runs, its restriction to static CGs leads to slightly worse performance.
On the other hand, FV-MCTS-MP can solve tasks even with $48$ agents successfully. 
Moreover, even on the eight agent problem, FV-MCTS-MP is much faster, taking on average approximately \SI{1}{\second} instead of \SI{40}{\second} for FV-MCTS-Var-El for the same tree search hyperparameters. 
\emph{FV-MCTS-MP scales to MMDP problem sizes that FV-MCTS-Var-El cannot even accommodate}. 

%% file: conclusion.tex
\section{Conclusion}
\label{sec:conclusion}
We introduced a scalable online planning algorithm for multi-agent MDPs with dynamic coordination graphs. Our approach, FV-MCTS-MP, uses Max-Plus for action coordination, in contrast to the previously introduced FV-MCTS with Variable Elimination. Over the standard SysAdmin and the custom Multi-Drone Delivery domains, we demonstrated that FV-MCTS-MP performs as well as Var-El on static CGs, outperforms it significantly on dynamic CGs, and is far more computationally efficient (enabling online MMDP planning on previously intractable problems).

In the appendix, we discuss how our approach can easily extend to multi-agent POMDPs.
However, we still require a domain expert to pre-define the appropriate coordination graph for the problem. A predetermined CG can be particularly difficult with highly dynamic domains where the CG depends on the state.
More work is required towards learning the dynamic coordination graph itself via interaction with the model.
Similarly, extending ideas from Alpha-Zero~\citep{silver2018general,anthony2017thinking} would be particularly relevant for distilling the coordinated individual actions for the agents into decentralized policies~\citep{phan2019distributed} with FV-MCTS acting as a scalable policy improvement operator \citep{grill2020monte}.

%% file: appendix.tex
\onecolumn
\appendix
\appendixpage

\section{Effect of Hyperparameters}
\subsection{SysAdmin}
We keep the number of agents fixed to $32$ for the following experiments.

\begin{figure*}
    \centering
    \begin{subfigure}{\textwidth}
    \centering
    \begin{subfigure}{0.32\columnwidth}
    \centering
    \includegraphics[width=\columnwidth]{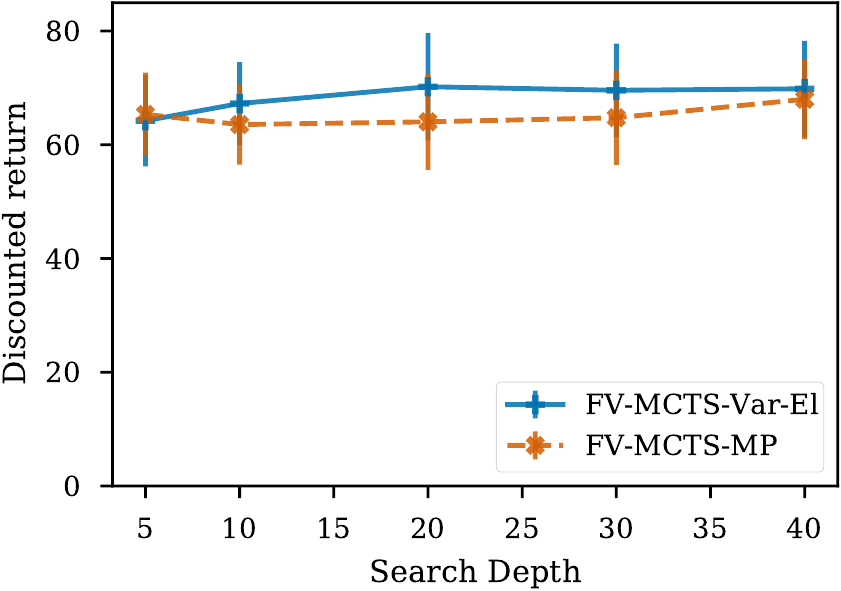}
    \end{subfigure}
    \begin{subfigure}{0.32\columnwidth}
    \centering
    \includegraphics[width=\columnwidth]{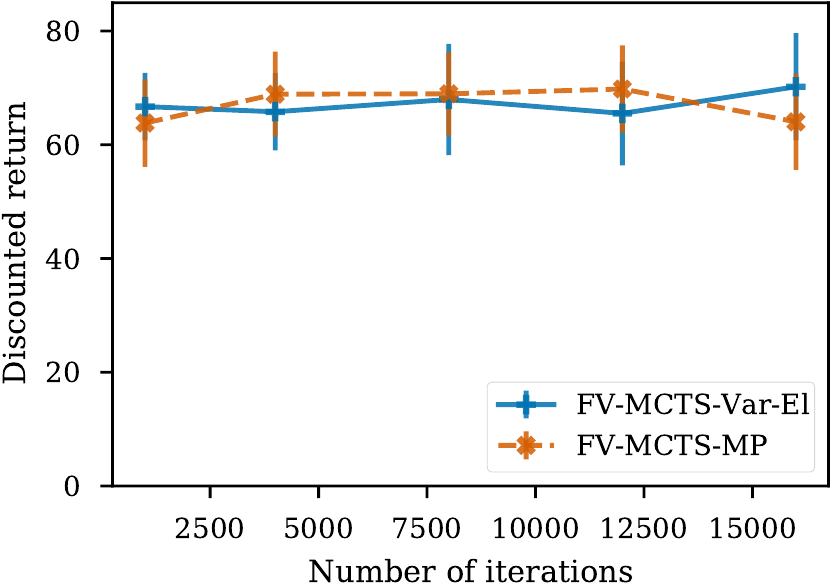}
    \end{subfigure}
    \begin{subfigure}{0.32\columnwidth}
    \centering
    \includegraphics[width=\columnwidth]{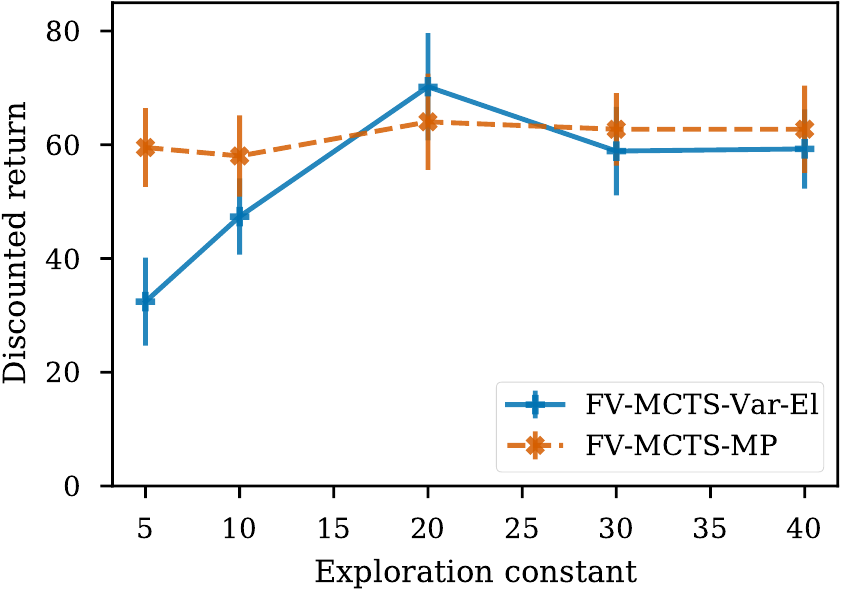}
    \end{subfigure}
    \caption{Star Sysadmin}
    \end{subfigure}
    \begin{subfigure}{\textwidth}
    \centering
    \begin{subfigure}{0.32\columnwidth}
    \centering
    \includegraphics[width=\columnwidth]{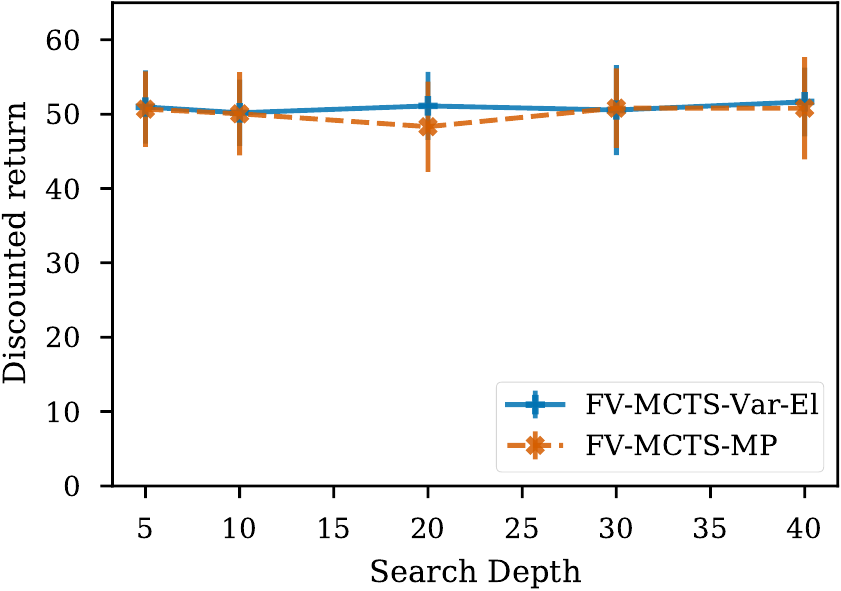}
    \end{subfigure}
    \begin{subfigure}{0.32\columnwidth}
    \centering
    \includegraphics[width=\columnwidth]{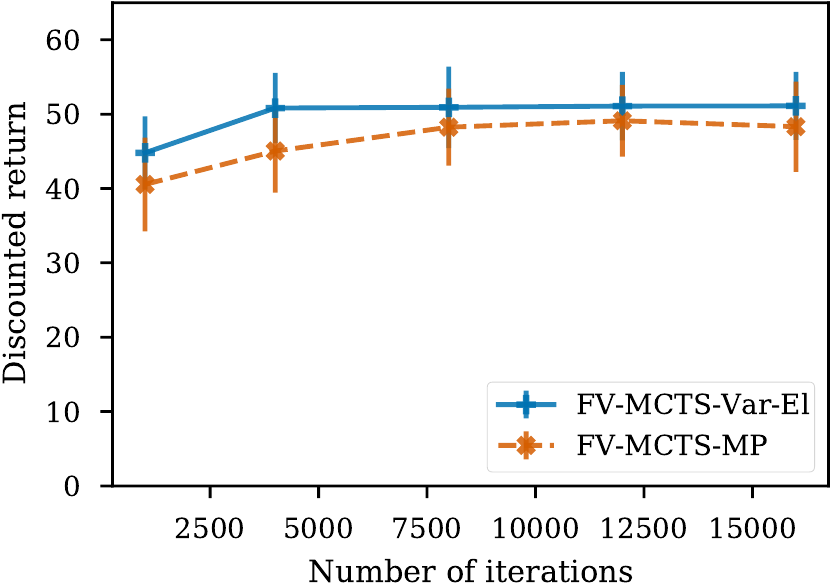}
    \end{subfigure}
    \begin{subfigure}{0.32\columnwidth}
    \centering
    \includegraphics[width=\columnwidth]{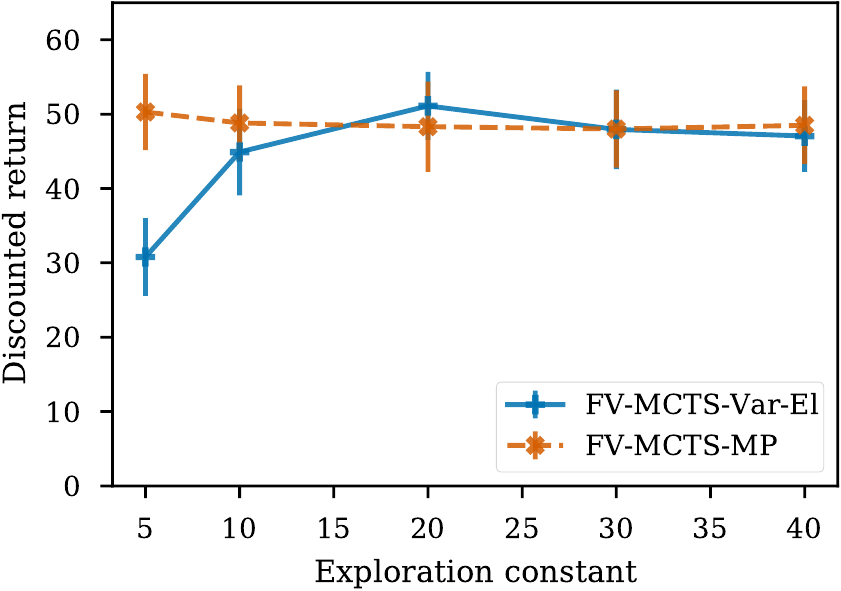}
    \end{subfigure}
    \caption{Ring Sysadmin}
    \end{subfigure}
    \caption{Effect of hyperparameters. $32$ agents.}
    \label{fig:hyperparam_fig}
\end{figure*}

\textbf{Exploration constant $c$}: Keeping number of Monte Carlo iterations fixed at $16000$ and exploration depth fixed at $20$, we varied the exploration constant between $5$ and $40$.
As can be seen from \Cref{fig:hyperparam_fig}, the difference is performance is not significant over the range.

\textbf{Exploration depth $d$}: Keeping the number of Monte Carlo iterations fixed at $16000$ and exploration constant fixed at $20$, we varied the tree search exploration depth.
We found that low exploration can adversely affect FV-MCTS-Var-El, but otherwise does not effect the performance very much.

\textbf{Number of Monte Carlo rollouts $n$}: Keeping the exploration constant fixed at $20$ and exploration depth fixed at $20$, we varied the number of Monte Carlo itertations used by the algorithms.
Low values adversely effected the performance a little.

\section{Algorithm Details}
For completeness we also also describe the FV-MCTS Var-El approach in~\Cref{algo:fvmctsvarel}. Note the difference in structure of tracked statistics.

\begin{algorithm}
  \caption{FV-MCTS-Var-El}
  \label{algo:fvmctsvarel}
  \begin{algorithmic}[1]
    \Require{{time limit, exploration depth, exploration constant}}
    \Statex
    \State $n_e, Q_e$ \Comment{Component statistics}
    \State
    \Function{FV-MCTS-Var-El}{$s$, depth}
    \While{time limit not reached}
    \State{\Call{Simulate}{$s$, depth}}
    \EndWhile
    \State{$\mathbf{a}^* \gets$ \Call{Var-El}{}}
    \State{\Return{$\mathbf{a}^*$}}
    \EndFunction
    \Function{Simulate}{$s$, depth}
    \If{$\gamma^{\mathrm{depth}} < \epsilon$}
    \State{\Return $0$}
    \EndIf
    \State{$a\gets$ \Call{Var-El}{exploration constant}}
    \State{$s', r \sim T(s,{a}), R(s, a)$}
    \State{$q \gets r + \gamma$ \Call{Simulate}{$s'$, depth $- 1$}}
    \State{\Call{UpdateStats}{$s, a, q$}}    
    \EndFunction
    
    \Function{UpdateStats}{$s, a, q$}
    \For{every component $e \in \mathcal{G}(s)$}
    \State{$n_{e}(s, a_e) \mathrel{{+}{=}} 1$}
    \State{$Q_{e}(s, a_e) \mathrel{{+}{=}} \frac{q - Q_{e}(s, a_e)}{n_{e}(s, a_e)} $}
    \EndFor
    \EndFunction    
  \end{algorithmic}
\end{algorithm}

\section{On the application to Multi-Agent POMDPs}

Strictly speaking, the prior work of~\citet{Amato2014-io} plans for Multi-Agent POMDPs (rather than MDPs), where the agents receive noisy observations of the true underlying state.
Given that our work is for MMDPs, we briefly comment on how the key challenges for both of us appear in the MMDP formulation itself and on how our approach could be applied directly to an MPOMDP. That is why we chose to use the MMDP for our implementation and evaluations rather than introducing the extraneous confound of a partially observable setting.

First, the prior work focuses almost entirely on the challenge of the exploding action space in a Monte Carlo Tree Search planning algorithm, which is just as pertinent for an MMDP as it is for an MPOMDP.  To handle partial observability they use POMCP, a partially observable extension of MCTS (specifically, Upper Confidence Trees) in which particle filters track and update the belief state with new observations~\cite{DBLP:conf/nips/SilverV10}. Our approach could also be used straightforwardly with POMCP and applied to MPOMDPs. 

Second, the only segment of the prior work that tries to address the unique challenge of an MPOMDP is a variant of their algorithm that splits joint observation histories into local histories and distributes them over the factors. In MMDPs there is no such thing as an observation history, since the current system state is fully observed; thus there is no equivalent variant to implement for the evaluation. Furthermore, for the MPOMDP case we could again incorporate our approach into their factorized trees variant directly, since our modifications would be to the action selection and exploration steps.

%% file: main.bbl

\begin{thebibliography}{43}


\ifx \showCODEN    \undefined \def \showCODEN     #1{\unskip}     \fi
\ifx \showDOI      \undefined \def \showDOI       #1{#1}\fi
\ifx \showISBNx    \undefined \def \showISBNx     #1{\unskip}     \fi
\ifx \showISBNxiii \undefined \def \showISBNxiii  #1{\unskip}     \fi
\ifx \showISSN     \undefined \def \showISSN      #1{\unskip}     \fi
\ifx \showLCCN     \undefined \def \showLCCN      #1{\unskip}     \fi
\ifx \shownote     \undefined \def \shownote      #1{#1}          \fi
\ifx \showarticletitle \undefined \def \showarticletitle #1{#1}   \fi
\ifx \showURL      \undefined \def \showURL       {\relax}        \fi
\providecommand\bibfield[2]{#2}
\providecommand\bibinfo[2]{#2}
\providecommand\natexlab[1]{#1}
\providecommand\showeprint[2][]{arXiv:#2}

\bibitem[\protect\citeauthoryear{Amato and Oliehoek}{Amato and
  Oliehoek}{2014}]%
        {Amato2014-io}
\bibfield{author}{\bibinfo{person}{Christopher Amato} {and}
  \bibinfo{person}{Frans~A Oliehoek}.} \bibinfo{year}{2014}\natexlab{}.
\newblock \showarticletitle{Scalable {P}lanning and {L}earning for {M}ultiagent
  {POMDPs}: Extended Version}.
\newblock \bibinfo{journal}{\emph{arXiv preprint arXiv:1404.1140}}
  (\bibinfo{year}{2014}).
\newblock
\showeprint[arxiv]{1404.1140}


\bibitem[\protect\citeauthoryear{Anthony, Tian, and Barber}{Anthony
  et~al\mbox{.}}{2017}]%
        {anthony2017thinking}
\bibfield{author}{\bibinfo{person}{Thomas Anthony}, \bibinfo{person}{Zheng
  Tian}, {and} \bibinfo{person}{David Barber}.}
  \bibinfo{year}{2017}\natexlab{}.
\newblock \showarticletitle{Thinking {F}ast and {S}low with {D}eep {L}earning
  and {T}ree {S}earch}. In \bibinfo{booktitle}{\emph{Advances in Neural
  Information Processing Systems (NeurIPS)}}. \bibinfo{pages}{5360--5370}.
\newblock


\bibitem[\protect\citeauthoryear{Bertsekas}{Bertsekas}{2005}]%
        {bertsekas2005dynamic}
\bibfield{author}{\bibinfo{person}{Dimitri~P Bertsekas}.}
  \bibinfo{year}{2005}\natexlab{}.
\newblock \bibinfo{booktitle}{\emph{Dynamic {P}rogramming and {O}ptimal
  {C}ontrol}}. Vol.~\bibinfo{volume}{1}.
\newblock \bibinfo{publisher}{Athena Scientific Belmont, MA}.
\newblock


\bibitem[\protect\citeauthoryear{Best, Cliff, Patten, Mettu, and Fitch}{Best
  et~al\mbox{.}}{2019}]%
        {Best2019-rr}
\bibfield{author}{\bibinfo{person}{Graeme Best}, \bibinfo{person}{Oliver~M
  Cliff}, \bibinfo{person}{Timothy Patten}, \bibinfo{person}{Ramgopal~R Mettu},
  {and} \bibinfo{person}{Robert Fitch}.} \bibinfo{year}{2019}\natexlab{}.
\newblock \showarticletitle{{Dec-MCTS}: {D}ecentralized {P}lanning for
  {M}ulti-{R}obot {A}ctive {P}erception}.
\newblock \bibinfo{journal}{\emph{International Journal of Robotics Research}}
  \bibinfo{volume}{38}, \bibinfo{number}{2-3} (\bibinfo{date}{March}
  \bibinfo{year}{2019}), \bibinfo{pages}{316--337}.
\newblock


\bibitem[\protect\citeauthoryear{Bezanson, Edelman, Karpinski, and
  Shah}{Bezanson et~al\mbox{.}}{2017}]%
        {bezanson2017julia}
\bibfield{author}{\bibinfo{person}{Jeff Bezanson}, \bibinfo{person}{Alan
  Edelman}, \bibinfo{person}{Stefan Karpinski}, {and} \bibinfo{person}{Viral~B
  Shah}.} \bibinfo{year}{2017}\natexlab{}.
\newblock \showarticletitle{Julia: A {F}resh {A}pproach to {N}umerical
  {C}omputing}.
\newblock \bibinfo{journal}{\emph{SIAM Rev.}} \bibinfo{volume}{59},
  \bibinfo{number}{1} (\bibinfo{year}{2017}), \bibinfo{pages}{65--98}.
\newblock


\bibitem[\protect\citeauthoryear{B{\"o}hmer, Kurin, and Whiteson}{B{\"o}hmer
  et~al\mbox{.}}{2019}]%
        {Bohmer2019-zv}
\bibfield{author}{\bibinfo{person}{Wendelin B{\"o}hmer},
  \bibinfo{person}{Vitaly Kurin}, {and} \bibinfo{person}{Shimon Whiteson}.}
  \bibinfo{year}{2019}\natexlab{}.
\newblock \showarticletitle{Deep {C}oordination {G}raphs}.
\newblock \bibinfo{journal}{\emph{arXiv preprint arXiv:1910.00091}}
  (\bibinfo{year}{2019}).
\newblock


\bibitem[\protect\citeauthoryear{Boutilier}{Boutilier}{1996}]%
        {boutilier1996planning}
\bibfield{author}{\bibinfo{person}{Craig Boutilier}.}
  \bibinfo{year}{1996}\natexlab{}.
\newblock \showarticletitle{Planning, {L}earning and {C}oordination in
  {M}ulti-{A}gent {D}ecision {P}rocesses}. In
  \bibinfo{booktitle}{\emph{Proceedings of the Sixth Conference on Theoretical
  Aspects of Rationality and Knowledge}}. Morgan Kaufmann Publishers Inc.,
  \bibinfo{pages}{195--210}.
\newblock


\bibitem[\protect\citeauthoryear{Browne, Powley, Whitehouse, Lucas, Cowling,
  Rohlfshagen, Tavener, Liebana, Samothrakis, and Colton}{Browne
  et~al\mbox{.}}{2012}]%
        {DBLP:journals/tciaig/BrownePWLCRTPSC12}
\bibfield{author}{\bibinfo{person}{Cameron Browne},
  \bibinfo{person}{Edward~Jack Powley}, \bibinfo{person}{Daniel Whitehouse},
  \bibinfo{person}{Simon~M. Lucas}, \bibinfo{person}{Peter~I. Cowling},
  \bibinfo{person}{Philipp Rohlfshagen}, \bibinfo{person}{Stephen Tavener},
  \bibinfo{person}{Diego~Perez Liebana}, \bibinfo{person}{Spyridon
  Samothrakis}, {and} \bibinfo{person}{Simon Colton}.}
  \bibinfo{year}{2012}\natexlab{}.
\newblock \showarticletitle{A {S}urvey of {M}onte {C}arlo {T}ree {S}earch
  {M}ethods}.
\newblock \bibinfo{journal}{\emph{{IEEE} Transactions on Computational
  Intelligence and AI in games}} \bibinfo{volume}{4}, \bibinfo{number}{1}
  (\bibinfo{year}{2012}), \bibinfo{pages}{1--43}.
\newblock


\bibitem[\protect\citeauthoryear{Chaslot, Winands, van~den Herik, Uiterwijk,
  and Bouzy}{Chaslot et~al\mbox{.}}{2008}]%
        {chaslot2008progressive}
\bibfield{author}{\bibinfo{person}{Guillaume M J-B Chaslot},
  \bibinfo{person}{Mark~HM Winands}, \bibinfo{person}{H~Jaap van~den Herik},
  \bibinfo{person}{Jos~WHM Uiterwijk}, {and} \bibinfo{person}{Bruno Bouzy}.}
  \bibinfo{year}{2008}\natexlab{}.
\newblock \showarticletitle{Progressive {S}trategies for {M}onte-{C}arlo {T}ree
  {S}earch}.
\newblock \bibinfo{journal}{\emph{New Mathematics and Natural Computation}}
  \bibinfo{volume}{4}, \bibinfo{number}{3} (\bibinfo{year}{2008}),
  \bibinfo{pages}{343--357}.
\newblock


\bibitem[\protect\citeauthoryear{Choudhury, Knickerbocker, and
  Kochenderfer}{Choudhury et~al\mbox{.}}{2019}]%
        {choudhury2019dynamic}
\bibfield{author}{\bibinfo{person}{Shushman Choudhury},
  \bibinfo{person}{Jacob~P. Knickerbocker}, {and} \bibinfo{person}{Mykel~J.
  Kochenderfer}.} \bibinfo{year}{2019}\natexlab{}.
\newblock \showarticletitle{Dynamic {R}eal-time {M}ultimodal {R}outing with
  {H}ierarchical {H}ybrid {P}lanning}. In \bibinfo{booktitle}{\emph{IEEE
  Intelligent Vehicles Symposium (IV)}}. \bibinfo{pages}{2397--2404}.
\newblock


\bibitem[\protect\citeauthoryear{Choudhury, Solovey, Kochenderfer, and
  Pavone}{Choudhury et~al\mbox{.}}{2020}]%
        {ChoudhurySoloveyETAL2020}
\bibfield{author}{\bibinfo{person}{Shushman Choudhury}, \bibinfo{person}{Kiril
  Solovey}, \bibinfo{person}{Mykel~J. Kochenderfer}, {and}
  \bibinfo{person}{Marco Pavone}.} \bibinfo{year}{2020}\natexlab{}.
\newblock \showarticletitle{Efficient {L}arge-{S}cale {M}ulti-{D}rone
  {D}elivery Using {T}ransit {N}etworks}. In \bibinfo{booktitle}{\emph{IEEE
  International Conference on Robotics and Automation (ICRA)}}.
\newblock


\bibitem[\protect\citeauthoryear{Dechter}{Dechter}{1999}]%
        {DBLP:journals/ai/Dechter99}
\bibfield{author}{\bibinfo{person}{Rina Dechter}.}
  \bibinfo{year}{1999}\natexlab{}.
\newblock \showarticletitle{Bucket {E}limination: {A} {U}nifying {F}ramework
  for {R}easoning}.
\newblock \bibinfo{journal}{\emph{Artificial Intelligence}}
  \bibinfo{volume}{113}, \bibinfo{number}{1-2} (\bibinfo{year}{1999}),
  \bibinfo{pages}{41--85}.
\newblock
\urldef\tempurl%
\url{https://doi.org/10.1016/S0004-3702(99)00059-4}
\showDOI{\tempurl}


\bibitem[\protect\citeauthoryear{Dorling, Heinrichs, Messier, and
  Magierowski}{Dorling et~al\mbox{.}}{2016}]%
        {dorling2016vehicle}
\bibfield{author}{\bibinfo{person}{Kevin Dorling}, \bibinfo{person}{Jordan
  Heinrichs}, \bibinfo{person}{Geoffrey~G Messier}, {and}
  \bibinfo{person}{Sebastian Magierowski}.} \bibinfo{year}{2016}\natexlab{}.
\newblock \showarticletitle{Vehicle {R}outing {P}roblems for {D}rone
  {D}elivery}.
\newblock \bibinfo{journal}{\emph{IEEE Transactions on Systems, Man, and
  Cybernetics: Systems}} \bibinfo{volume}{47}, \bibinfo{number}{1}
  (\bibinfo{year}{2016}), \bibinfo{pages}{70--85}.
\newblock


\bibitem[\protect\citeauthoryear{Grill, Altch{\'e}, Tang, Hubert, Valko,
  Antonoglou, and Munos}{Grill et~al\mbox{.}}{2020}]%
        {grill2020monte}
\bibfield{author}{\bibinfo{person}{Jean-Bastien Grill},
  \bibinfo{person}{Florent Altch{\'e}}, \bibinfo{person}{Yunhao Tang},
  \bibinfo{person}{Thomas Hubert}, \bibinfo{person}{Michal Valko},
  \bibinfo{person}{Ioannis Antonoglou}, {and} \bibinfo{person}{R{\'e}mi
  Munos}.} \bibinfo{year}{2020}\natexlab{}.
\newblock \showarticletitle{{M}onte-{C}arlo {T}ree {S}earch as {R}egularized
  {P}olicy {O}ptimization}.
\newblock \bibinfo{journal}{\emph{arXiv preprint arXiv:2007.12509}}
  (\bibinfo{year}{2020}).
\newblock


\bibitem[\protect\citeauthoryear{Guestrin, Koller, and Parr}{Guestrin
  et~al\mbox{.}}{2002}]%
        {Guestrin2002-il}
\bibfield{author}{\bibinfo{person}{Carlos Guestrin}, \bibinfo{person}{Daphne
  Koller}, {and} \bibinfo{person}{Ronald Parr}.}
  \bibinfo{year}{2002}\natexlab{}.
\newblock \showarticletitle{Multiagent {P}lanning with {F}actored {MDPs}}.
\newblock In \bibinfo{booktitle}{\emph{Advances in Neural Information
  Processing Systems}}, \bibfield{editor}{\bibinfo{person}{T~G Dietterich},
  \bibinfo{person}{S~Becker}, {and} \bibinfo{person}{Z~Ghahramani}} (Eds.).
  \bibinfo{publisher}{MIT Press}, \bibinfo{pages}{1523--1530}.
\newblock


\bibitem[\protect\citeauthoryear{Guestrin, Koller, Parr, and
  Venkataraman}{Guestrin et~al\mbox{.}}{2003}]%
        {Guestrin2003}
\bibfield{author}{\bibinfo{person}{C. Guestrin}, \bibinfo{person}{D. Koller},
  \bibinfo{person}{R. Parr}, {and} \bibinfo{person}{S. Venkataraman}.}
  \bibinfo{year}{2003}\natexlab{}.
\newblock \showarticletitle{Efficient {S}olution {A}lgorithms for {F}actored
  {MDPs}}.
\newblock \bibinfo{journal}{\emph{Journal of Artificial Intelligence Research}}
   \bibinfo{volume}{19} (\bibinfo{date}{Oct.} \bibinfo{year}{2003}),
  \bibinfo{pages}{399--468}.
\newblock
\urldef\tempurl%
\url{https://doi.org/10.1613/jair.1000}
\showDOI{\tempurl}


\bibitem[\protect\citeauthoryear{Gupta, Egorov, and Kochenderfer}{Gupta
  et~al\mbox{.}}{2017}]%
        {gupta2017cooperative}
\bibfield{author}{\bibinfo{person}{Jayesh~K Gupta}, \bibinfo{person}{Maxim
  Egorov}, {and} \bibinfo{person}{Mykel~J. Kochenderfer}.}
  \bibinfo{year}{2017}\natexlab{}.
\newblock \showarticletitle{Cooperative {M}ulti-{A}gent {C}ontrol using {D}eep
  {R}einforcement {L}earning}. In \bibinfo{booktitle}{\emph{International
  Conference on Autonomous Agents and Multiagent Systems (AAMAS)}}. Springer,
  \bibinfo{pages}{66--83}.
\newblock


\bibitem[\protect\citeauthoryear{Kochenderfer}{Kochenderfer}{2015}]%
        {kochenderfer2015decision}
\bibfield{author}{\bibinfo{person}{Mykel~J. Kochenderfer}.}
  \bibinfo{year}{2015}\natexlab{}.
\newblock \bibinfo{booktitle}{\emph{Decision {Making} under {Uncertainty}:
  {Theory} and {Application}}}.
\newblock \bibinfo{publisher}{MIT Press}.
\newblock


\bibitem[\protect\citeauthoryear{Kocsis and Szepesv{\'{a}}ri}{Kocsis and
  Szepesv{\'{a}}ri}{2006}]%
        {DBLP:conf/ecml/KocsisS06}
\bibfield{author}{\bibinfo{person}{Levente Kocsis} {and} \bibinfo{person}{Csaba
  Szepesv{\'{a}}ri}.} \bibinfo{year}{2006}\natexlab{}.
\newblock \showarticletitle{Bandit Based Monte-Carlo Planning}. In
  \bibinfo{booktitle}{\emph{European Conference on Machine Learning (ECML)}},
  Vol.~\bibinfo{volume}{4212}. \bibinfo{publisher}{Springer},
  \bibinfo{pages}{282--293}.
\newblock
\urldef\tempurl%
\url{https://doi.org/10.1007/11871842\_29}
\showDOI{\tempurl}


\bibitem[\protect\citeauthoryear{Kok, Hoen, Bakker, and Vlassis}{Kok
  et~al\mbox{.}}{2005}]%
        {kok2005utile}
\bibfield{author}{\bibinfo{person}{Jelle~R Kok}, \bibinfo{person}{Eter~Jan
  Hoen}, \bibinfo{person}{Bram Bakker}, {and} \bibinfo{person}{Nikos Vlassis}.}
  \bibinfo{year}{2005}\natexlab{}.
\newblock \showarticletitle{Utile {C}oordination: {L}earning
  {I}nterdependencies {A}mong {C}ooperative {A}gents}. In
  \bibinfo{booktitle}{\emph{EEE Symposium on Computational Intelligence and
  Games, Colchester, Essex}}. \bibinfo{pages}{29--36}.
\newblock


\bibitem[\protect\citeauthoryear{Kok and Vlassis}{Kok and Vlassis}{2004}]%
        {kok2004sparse}
\bibfield{author}{\bibinfo{person}{Jelle~R Kok} {and} \bibinfo{person}{Nikos
  Vlassis}.} \bibinfo{year}{2004}\natexlab{}.
\newblock \showarticletitle{Sparse {C}ooperative {Q}-{L}earning}. In
  \bibinfo{booktitle}{\emph{International Conference on Machine Learning
  (ICML)}}. ACM, \bibinfo{pages}{61}.
\newblock


\bibitem[\protect\citeauthoryear{Kok and Vlassis}{Kok and Vlassis}{2005}]%
        {DBLP:conf/bnaic/KokV05}
\bibfield{author}{\bibinfo{person}{Jelle~R. Kok} {and}
  \bibinfo{person}{Nikos~A. Vlassis}.} \bibinfo{year}{2005}\natexlab{}.
\newblock \showarticletitle{Using the {M}ax-{P}lus {A}lgorithm for
  {M}ulti-{A}gent {D}ecision {M}aking in {C}oordination {G}raphs}. In
  \bibinfo{booktitle}{\emph{Proceedings of the Seventeenth Belgium-Netherlands
  Conference on Artificial Intelligence {(BNAIC)}}}. \bibinfo{pages}{359--360}.
\newblock


\bibitem[\protect\citeauthoryear{Kuyer, Whiteson, Bakker, and Vlassis}{Kuyer
  et~al\mbox{.}}{2008}]%
        {kuyer2008multiagent}
\bibfield{author}{\bibinfo{person}{Lior Kuyer}, \bibinfo{person}{Shimon
  Whiteson}, \bibinfo{person}{Bram Bakker}, {and} \bibinfo{person}{Nikos
  Vlassis}.} \bibinfo{year}{2008}\natexlab{}.
\newblock \showarticletitle{Multiagent {R}einforcement {L}earning for {U}rban
  {T}raffic {C}ontrol using {C}oordination {G}raphs}. In
  \bibinfo{booktitle}{\emph{Joint European Conference on Machine Learning and
  Knowledge Discovery in Databases}}. Springer, \bibinfo{pages}{656--671}.
\newblock


\bibitem[\protect\citeauthoryear{Lee}{Lee}{2017}]%
        {DBLP:conf/syscon/Lee17}
\bibfield{author}{\bibinfo{person}{Jaihyun Lee}.}
  \bibinfo{year}{2017}\natexlab{}.
\newblock \showarticletitle{Optimization of a {M}odular {D}rone {D}elivery
  {S}ystem}. In \bibinfo{booktitle}{\emph{IEEE International Systems
  Conference}}. \bibinfo{pages}{1--8}.
\newblock
\urldef\tempurl%
\url{https://doi.org/10.1109/SYSCON.2017.7934790}
\showDOI{\tempurl}


\bibitem[\protect\citeauthoryear{Li, Gupta, Morales, Allen, and
  Kochenderfer}{Li et~al\mbox{.}}{2021}]%
        {li2020deep}
\bibfield{author}{\bibinfo{person}{Sheng Li}, \bibinfo{person}{Jayesh~K Gupta},
  \bibinfo{person}{Peter Morales}, \bibinfo{person}{Ross Allen}, {and}
  \bibinfo{person}{Mykel~J. Kochenderfer}.} \bibinfo{year}{2021}\natexlab{}.
\newblock \showarticletitle{Deep {I}mplicit {C}oordination {G}raphs for
  {M}ulti-agent {R}einforcement {L}earning}. In
  \bibinfo{booktitle}{\emph{International Conference on Autonomous Agents and
  Multiagent Systems (AAMAS)}}.
\newblock


\bibitem[\protect\citeauthoryear{Matignon, Laurent, and Le~Fort-Piat}{Matignon
  et~al\mbox{.}}{2012}]%
        {matignon2012}
\bibfield{author}{\bibinfo{person}{Laetitia Matignon},
  \bibinfo{person}{Guillaume~J. Laurent}, {and} \bibinfo{person}{Nadine
  Le~Fort-Piat}.} \bibinfo{year}{2012}\natexlab{}.
\newblock \showarticletitle{Independent {R}einforcement {L}earners in
  {C}ooperative {M}arkov {G}ames: a {S}urvey {R}egarding {C}oordination
  {P}roblems}.
\newblock \bibinfo{journal}{\emph{The Knowledge Engineering Review}}
  \bibinfo{volume}{27}, \bibinfo{number}{1} (\bibinfo{year}{2012}),
  \bibinfo{pages}{1–31}.
\newblock
\urldef\tempurl%
\url{https://doi.org/10.1017/S0269888912000057}
\showDOI{\tempurl}


\bibitem[\protect\citeauthoryear{Murphy, Weiss, and Jordan}{Murphy
  et~al\mbox{.}}{1999}]%
        {DBLP:conf/uai/MurphyWJ99}
\bibfield{author}{\bibinfo{person}{Kevin~P. Murphy}, \bibinfo{person}{Yair
  Weiss}, {and} \bibinfo{person}{Michael~I. Jordan}.}
  \bibinfo{year}{1999}\natexlab{}.
\newblock \showarticletitle{Loopy {B}elief {P}ropagation for {A}pproximate
  {I}nference: {A}n {E}mpirical {S}tudy}. In
  \bibinfo{booktitle}{\emph{Conference on Uncertainty in Artificial
  Intelligence (UAI)}}. \bibinfo{publisher}{Morgan Kaufmann},
  \bibinfo{pages}{467--475}.
\newblock


\bibitem[\protect\citeauthoryear{Nijssen and Winands}{Nijssen and
  Winands}{2011}]%
        {nijssen2011}
\bibfield{author}{\bibinfo{person}{J.~A.~M. Nijssen} {and}
  \bibinfo{person}{Mark H.~M. Winands}.} \bibinfo{year}{2011}\natexlab{}.
\newblock \showarticletitle{{E}nhancements for {M}ulti-{P}layer {M}onte-{C}arlo
  {T}ree {S}earch}. In \bibinfo{booktitle}{\emph{Computers and Games}}.
\newblock


\bibitem[\protect\citeauthoryear{Oh, Park, and Ahn}{Oh et~al\mbox{.}}{2015}]%
        {oh2015survey}
\bibfield{author}{\bibinfo{person}{Kwang-Kyo Oh}, \bibinfo{person}{Myoung-Chul
  Park}, {and} \bibinfo{person}{Hyo-Sung Ahn}.}
  \bibinfo{year}{2015}\natexlab{}.
\newblock \showarticletitle{A {S}urvey of {M}ulti-{A}gent {F}ormation
  {C}ontrol}.
\newblock \bibinfo{journal}{\emph{Automatica}}  \bibinfo{volume}{53}
  (\bibinfo{year}{2015}), \bibinfo{pages}{424--440}.
\newblock


\bibitem[\protect\citeauthoryear{Oliehoek, Spaan, Whiteson, and
  Vlassis}{Oliehoek et~al\mbox{.}}{2008}]%
        {DBLP:conf/atal/OliehoekSWV08}
\bibfield{author}{\bibinfo{person}{Frans~A. Oliehoek},
  \bibinfo{person}{Matthijs T.~J. Spaan}, \bibinfo{person}{Shimon Whiteson},
  {and} \bibinfo{person}{Nikos~A. Vlassis}.} \bibinfo{year}{2008}\natexlab{}.
\newblock \showarticletitle{Exploiting {L}ocality of {I}nteraction in
  {F}actored {D}ec-{POMDP}s}. In \bibinfo{booktitle}{\emph{International
  Conference on Autonomous Agents and Multiagent Systems (AAMAS)}}.
  \bibinfo{publisher}{International Foundation for Autonomous Agents and
  Multiagent Systems}, \bibinfo{pages}{517--524}.
\newblock


\bibitem[\protect\citeauthoryear{Pearl}{Pearl}{1989}]%
        {DBLP:books/daglib/Pearl89}
\bibfield{author}{\bibinfo{person}{Judea Pearl}.}
  \bibinfo{year}{1989}\natexlab{}.
\newblock \bibinfo{booktitle}{\emph{Probabilistic {R}easoning in {I}ntelligent
  {S}ystems: {N}etworks of {P}lausible {I}nference}}.
\newblock \bibinfo{publisher}{Morgan Kaufmann}.
\newblock


\bibitem[\protect\citeauthoryear{Phan, Schmid, Belzner, Gabor, Feld, and
  Linnhoff-Popien}{Phan et~al\mbox{.}}{2019}]%
        {phan2019distributed}
\bibfield{author}{\bibinfo{person}{Thomy Phan}, \bibinfo{person}{Kyrill
  Schmid}, \bibinfo{person}{Lenz Belzner}, \bibinfo{person}{Thomas Gabor},
  \bibinfo{person}{Sebastian Feld}, {and} \bibinfo{person}{Claudia
  Linnhoff-Popien}.} \bibinfo{year}{2019}\natexlab{}.
\newblock \showarticletitle{Distributed {P}olicy {I}teration for {S}calable
  {A}pproximation of {C}ooperative {M}ulti-{A}gent {P}olicies}. In
  \bibinfo{booktitle}{\emph{International Conference on Autonomous Agents and
  Multiagent Systems (AAMAS)}}. \bibinfo{pages}{2162--2164}.
\newblock


\bibitem[\protect\citeauthoryear{Pynadath and Tambe}{Pynadath and
  Tambe}{2002}]%
        {DBLP:journals/jair/PynadathT02}
\bibfield{author}{\bibinfo{person}{David~V. Pynadath} {and}
  \bibinfo{person}{Milind Tambe}.} \bibinfo{year}{2002}\natexlab{}.
\newblock \showarticletitle{The {C}ommunicative {M}ultiagent {T}eam {D}ecision
  {P}roblem: {A}nalyzing {T}eamwork {T}heories and {M}odels}.
\newblock \bibinfo{journal}{\emph{Journal of Artificial Intelligence Research}}
   \bibinfo{volume}{16} (\bibinfo{year}{2002}), \bibinfo{pages}{389--423}.
\newblock
\urldef\tempurl%
\url{https://doi.org/10.1613/jair.1024}
\showDOI{\tempurl}


\bibitem[\protect\citeauthoryear{Rashid, Samvelyan, Schroeder, Farquhar,
  Foerster, and Whiteson}{Rashid et~al\mbox{.}}{2018}]%
        {rashid2018qmix}
\bibfield{author}{\bibinfo{person}{Tabish Rashid}, \bibinfo{person}{Mikayel
  Samvelyan}, \bibinfo{person}{Christian Schroeder}, \bibinfo{person}{Gregory
  Farquhar}, \bibinfo{person}{Jakob Foerster}, {and} \bibinfo{person}{Shimon
  Whiteson}.} \bibinfo{year}{2018}\natexlab{}.
\newblock \showarticletitle{{QMIX}: {M}onotonic {V}alue {F}unction
  {F}actorisation for {D}eep {M}ulti-{A}gent {R}einforcement {L}earning}. In
  \bibinfo{booktitle}{\emph{International Conference on Machine Learning
  (ICML)}}. \bibinfo{pages}{4295--4304}.
\newblock


\bibitem[\protect\citeauthoryear{Silver, Hubert, Schrittwieser, Antonoglou,
  Lai, Guez, Lanctot, Sifre, Kumaran, Graepel, et~al\mbox{.}}{Silver
  et~al\mbox{.}}{2018}]%
        {silver2018general}
\bibfield{author}{\bibinfo{person}{David Silver}, \bibinfo{person}{Thomas
  Hubert}, \bibinfo{person}{Julian Schrittwieser}, \bibinfo{person}{Ioannis
  Antonoglou}, \bibinfo{person}{Matthew Lai}, \bibinfo{person}{Arthur Guez},
  \bibinfo{person}{Marc Lanctot}, \bibinfo{person}{Laurent Sifre},
  \bibinfo{person}{Dharshan Kumaran}, \bibinfo{person}{Thore Graepel},
  {et~al\mbox{.}}} \bibinfo{year}{2018}\natexlab{}.
\newblock \showarticletitle{A {G}eneral {R}einforcement {L}earning {A}lgorithm
  that {M}asters {C}hess, {S}hogi, and {G}o through {S}elf-{P}lay}.
\newblock \bibinfo{journal}{\emph{Science}} \bibinfo{volume}{362},
  \bibinfo{number}{6419} (\bibinfo{year}{2018}), \bibinfo{pages}{1140--1144}.
\newblock


\bibitem[\protect\citeauthoryear{Silver and Veness}{Silver and Veness}{2010}]%
        {DBLP:conf/nips/SilverV10}
\bibfield{author}{\bibinfo{person}{David Silver} {and} \bibinfo{person}{Joel
  Veness}.} \bibinfo{year}{2010}\natexlab{}.
\newblock \showarticletitle{Monte-{C}arlo {P}lanning in {L}arge {POMDP}s}. In
  \bibinfo{booktitle}{\emph{Advances in Neural Information Processing Systems
  (NeurIPS)}}. \bibinfo{publisher}{Curran Associates, Inc.},
  \bibinfo{pages}{2164--2172}.
\newblock


\bibitem[\protect\citeauthoryear{Sunehag, Lever, Gruslys, Czarnecki, Zambaldi,
  Jaderberg, Lanctot, Sonnerat, Leibo, Tuyls, et~al\mbox{.}}{Sunehag
  et~al\mbox{.}}{2018}]%
        {sunehag2018value}
\bibfield{author}{\bibinfo{person}{Peter Sunehag}, \bibinfo{person}{Guy Lever},
  \bibinfo{person}{Audrunas Gruslys}, \bibinfo{person}{Wojciech~Marian
  Czarnecki}, \bibinfo{person}{Vinicius Zambaldi}, \bibinfo{person}{Max
  Jaderberg}, \bibinfo{person}{Marc Lanctot}, \bibinfo{person}{Nicolas
  Sonnerat}, \bibinfo{person}{Joel~Z Leibo}, \bibinfo{person}{Karl Tuyls},
  {et~al\mbox{.}}} \bibinfo{year}{2018}\natexlab{}.
\newblock \showarticletitle{Value-{D}ecomposition {N}etworks for {C}ooperative
  {M}ulti-{A}gent {L}earning based on {T}eam {R}eward}. In
  \bibinfo{booktitle}{\emph{International Conference on Autonomous Agents and
  Multiagent Systems (AAMAS)}}. \bibinfo{pages}{2085--2087}.
\newblock


\bibitem[\protect\citeauthoryear{Sutton and Barto}{Sutton and Barto}{1998}]%
        {DBLP:books/lib/SuttonB98}
\bibfield{author}{\bibinfo{person}{Richard~S. Sutton} {and}
  \bibinfo{person}{Andrew~G. Barto}.} \bibinfo{year}{1998}\natexlab{}.
\newblock \bibinfo{booktitle}{\emph{Reinforcement Learning - An Introduction}}.
\newblock \bibinfo{publisher}{{MIT} Press}.
\newblock
\showISBNx{978-0-262-19398-6}


\bibitem[\protect\citeauthoryear{Vlassis, Elhorst, and Kok}{Vlassis
  et~al\mbox{.}}{2004}]%
        {Vlassis2004-da}
\bibfield{author}{\bibinfo{person}{N Vlassis}, \bibinfo{person}{R Elhorst},
  {and} \bibinfo{person}{J~R Kok}.} \bibinfo{year}{2004}\natexlab{}.
\newblock \showarticletitle{Anytime {A}lgorithms for {M}ultiagent {D}ecision
  {M}aking using {C}oordination {G}raphs}. In \bibinfo{booktitle}{\emph{{IEEE}
  International Conference on Systems, Man and Cybernetics ({IEEE} Cat.
  {No.04CH37583})}}, Vol.~\bibinfo{volume}{1}. \bibinfo{pages}{953--957 vol.1}.
\newblock


\bibitem[\protect\citeauthoryear{Wainwright, Jaakkola, and Willsky}{Wainwright
  et~al\mbox{.}}{2004}]%
        {DBLP:journals/sac/WainwrightJW04}
\bibfield{author}{\bibinfo{person}{Martin~J. Wainwright},
  \bibinfo{person}{Tommi~S. Jaakkola}, {and} \bibinfo{person}{Alan~S.
  Willsky}.} \bibinfo{year}{2004}\natexlab{}.
\newblock \showarticletitle{Tree {C}onsistency and {B}ounds on the
  {P}erformance of the {M}ax-{P}roduct {A}lgorithm and its {G}eneralizations}.
\newblock \bibinfo{journal}{\emph{Statistical Computing}} \bibinfo{volume}{14},
  \bibinfo{number}{2} (\bibinfo{year}{2004}), \bibinfo{pages}{143--166}.
\newblock
\urldef\tempurl%
\url{https://doi.org/10.1023/B:STCO.0000021412.33763.d5}
\showDOI{\tempurl}


\bibitem[\protect\citeauthoryear{Yu, Wang, Xu, Zhang, Ge, Ren, Sun, Chen, and
  Tan}{Yu et~al\mbox{.}}{2020}]%
        {yu2020dist}
\bibfield{author}{\bibinfo{person}{Chao Yu}, \bibinfo{person}{Xin Wang},
  \bibinfo{person}{Xin Xu}, \bibinfo{person}{Minjie Zhang},
  \bibinfo{person}{Hongwei Ge}, \bibinfo{person}{Jiankang Ren},
  \bibinfo{person}{Liang Sun}, \bibinfo{person}{Bingcai Chen}, {and}
  \bibinfo{person}{Guozhen Tan}.} \bibinfo{year}{2020}\natexlab{}.
\newblock \showarticletitle{Distributed {M}ultiagent {C}oordinated {L}earning
  for {A}utonomous {D}riving in {H}ighways {B}ased on {D}ynamic {C}oordination
  {G}raphs}.
\newblock \bibinfo{journal}{\emph{{IEEE} Trans. Intell. Transp. Syst.}}
  \bibinfo{volume}{21}, \bibinfo{number}{2} (\bibinfo{year}{2020}),
  \bibinfo{pages}{735--748}.
\newblock
\urldef\tempurl%
\url{https://doi.org/10.1109/TITS.2019.2893683}
\showDOI{\tempurl}


\bibitem[\protect\citeauthoryear{Zerbel and Yliniemi}{Zerbel and
  Yliniemi}{2019}]%
        {zerbel2019multiagent}
\bibfield{author}{\bibinfo{person}{Nicholas Zerbel} {and}
  \bibinfo{person}{Logan Yliniemi}.} \bibinfo{year}{2019}\natexlab{}.
\newblock \showarticletitle{Multiagent {M}onte {C}arlo {T}ree {S}earch}. In
  \bibinfo{booktitle}{\emph{International Conference on Autonomous Agents and
  Multiagent Systems (AAMAS)}}. \bibinfo{pages}{2309--2311}.
\newblock


\bibitem[\protect\citeauthoryear{Zykov}{Zykov}{1949}]%
        {zykov1949some}
\bibfield{author}{\bibinfo{person}{Alexander~Aleksandrovich Zykov}.}
  \bibinfo{year}{1949}\natexlab{}.
\newblock \showarticletitle{On {S}ome {P}roperties of {L}inear {C}omplexes}.
\newblock \bibinfo{journal}{\emph{Matematicheskii Sbornik}}
  \bibinfo{volume}{66}, \bibinfo{number}{2} (\bibinfo{year}{1949}),
  \bibinfo{pages}{163--188}.
\newblock


\end{thebibliography}
